\documentclass[journal,twocolumn,10pt]{IEEEtran}

\usepackage{floatflt} % text flows around figures
\usepackage{paralist} % compact and in-paragraph lists
\usepackage{algorithm} % package for pseudo code

\usepackage{graphicx}
\usepackage{amsmath}
\usepackage{amsthm}
\usepackage{amssymb}
\usepackage{listings}    %输入代码必备
\usepackage{xcolor}
\usepackage{indentfirst}
\usepackage{booktabs}
\usepackage{subfigure}
\usepackage{caption}
\usepackage{calligra}
\usepackage{bm}
\usepackage{fancyhdr}
\usepackage{float}
\usepackage{pifont}% http://ctan.org/pkg/pifont

\def\a{\mathbf{a}}

\def\x{\mathbf{x}}
\def\xc{\x^{\rm (c)}}

\def\y{\mathbf{y}}

\def\s{\mathbf{s}}

\def\vv{\mathbf{v}}

\def\N{\mathcal{N}}

\def\M{\mathcal{M}}

\def\one{{\bf 1}}

\newcommand{\mypar}[1]{{\bf #1.}}
\theoremstyle{definition}
\newtheorem{defn}{Definition}
\newtheorem{myLem}{Lemma}

\newtheorem{myThm}{Theorem}

\newtheorem{axiom}{Example}

\DeclareMathOperator{\LL}{L}

\DeclareMathOperator{\Adj}{A}

\DeclareMathOperator{\D}{D}

\DeclareMathOperator{\F}{F}

\DeclareMathOperator{\RR}{R}

\DeclareMathOperator{\Q}{Q}

\DeclareMathOperator{\Ss}{S}
\DeclareMathOperator{\Vm}{V}
\DeclareMathOperator{\X}{X}
\DeclareMathOperator{\Xc}{\X_{\rm c}}
\DeclareMathOperator{\Xo}{\X_{\rm o}}

\DeclareMathOperator{\W}{W}
\DeclareMathOperator{\Dd}{D}

\newcommand{\R}{\ensuremath{\mathbb{R}}}

\DeclareMathOperator{\Id}{I}

\usepackage{xr}
\begin{document}
\title{Fast Resampling of 3D Point Clouds via Graphs}
\author{Siheng~Chen,~\IEEEmembership{Student~Member,~IEEE}, Dong Tian,~\IEEEmembership{Senior Member,~IEEE},  Chen Feng,~\IEEEmembership{Member,~IEEE}, Anthony Vetro,~\IEEEmembership{Fellow,~IEEE}, Jelena~Kova\v{c}evi\'c,~\IEEEmembership{Fellow,~IEEE}% <-this % stops a space
  \thanks{}% <-this % stops a space
}
%% The paper headers
%\markboth{IEEE Trans. Signal Process., June 2016. In preparation.} {Chen \MakeLowercase{\textit{et al.}}: Resampling}
 \maketitle

%\tableofcontents

\begin{abstract}
To reduce cost in storing, processing and visualizing a large-scale point cloud,  we consider a randomized resampling strategy to select a representative subset of points while preserving application-dependent features. The proposed strategy is based on graphs, which can represent underlying surfaces and lend themselves well to efficient computation. We use a general feature-extraction operator to represent application-dependent features and propose a general reconstruction error to evaluate the quality of resampling. We obtain a general form of optimal resampling distribution by minimizing the reconstruction error. The proposed optimal resampling distribution is guaranteed to be shift, rotation and scale-invariant in the 3D space.  We next specify the  feature-extraction operator to be a graph filter and study specific resampling strategies based on all-pass, low-pass, high-pass graph filtering and graph filter banks. We finally apply the proposed methods to three applications:  large-scale visualization, accurate registration and robust shape modeling. The empirical performance validates the effectiveness and efficiency of the proposed resampling methods.
\end{abstract}


\section{Introduction}
With the recent development of 3D sensing technologies, 3D point clouds have become an important and practical representation of 3D objects and surrounding environments in many applications, such as virtual reality, mobile mapping, scanning of historical artifacts, 3D printing and digital elevation models~\cite{PCL}.
A large number of 3D points on an object's surface measured by a sensing device are called a~\emph{3D point cloud}. Other than 3D coordinates, a 3D point cloud may also comprise some attributes, such as color, temperature and texture.
Based on storage order and spatial connectivity between 3D points, there are two types of point clouds:~\emph{organized} point clouds and~\emph{unorganized} point clouds~\cite{HoppeDDMS:92}.
3D points collected by a camera-like 3D sensor or a 3D laser scanner are typically arranged on a grid, like pixels in an image; we call those point clouds organized.  For complex objects, we need to scan these objects from multiple view points and merge all collected points, which intermingles the indices of 3D points; we call those point clouds unorganized.  It is easier to process an organized point cloud than an unorganized point cloud as the underlying grid produces a natural spatial connectivity and reflects the order of sensing.  To make it general, we consider unorganized point clouds in this paper.

3D point cloud processing has become an important component in many 3D imaging and vision systems. It broadly includes compression~\cite{KammerlBRGBS:12, ZhangFL:14, ThanouCF:16,AnisCO:16}, visualization~\cite{OesterlingHJSH:11,PecnikMZ:13}, surface reconstruction~\cite{GregorskiHJ:00,GolovinskiyKF:09}, rendering~\cite{AlexaBCFLS:03,RydenKC:11}, editing~\cite{GuskovSS:99,WandBBFHJMSS:07} and feature extraction~\cite{DemarsinVVR:06, WeberHH:10,  DanielsHOS:07, ChoiTC:13, FengTK:14}.
A  challenge in 3D point cloud processing is how to handle a large number of incoming  3D points~\cite{ShahzadZ:15,LuoZ:15}. In many applications, such as digital documentation of historical buildings and terrain visualization, we need to store billions of incoming 3D points; additionally, real-time sensing systems generate millions of data points per second. A large-scale point cloud makes storage and subsequent processing inefficient.

\vspace{-4mm}
\begin{figure}[thb]
  \begin{center}
    \begin{tabular}{cc}
    \vspace{-4mm}
\includegraphics[width=0.42\columnwidth]{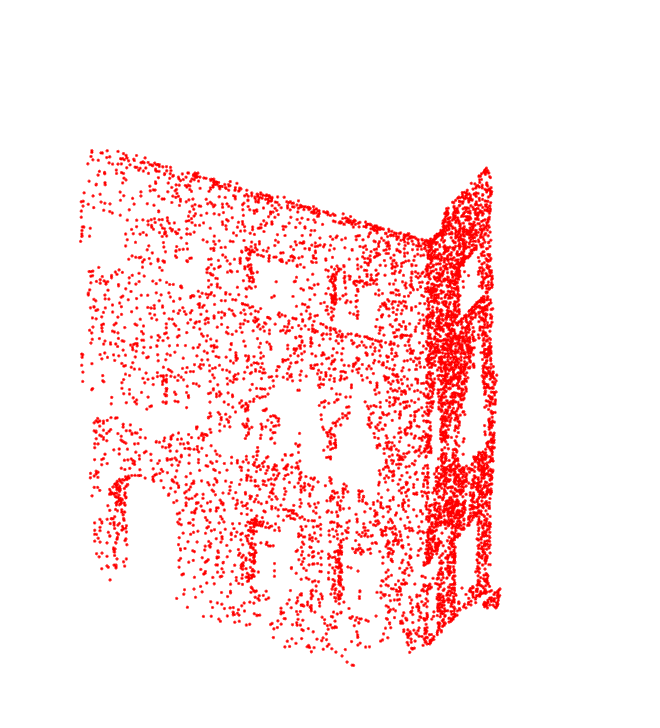}  
&
\includegraphics[width=0.42\columnwidth]{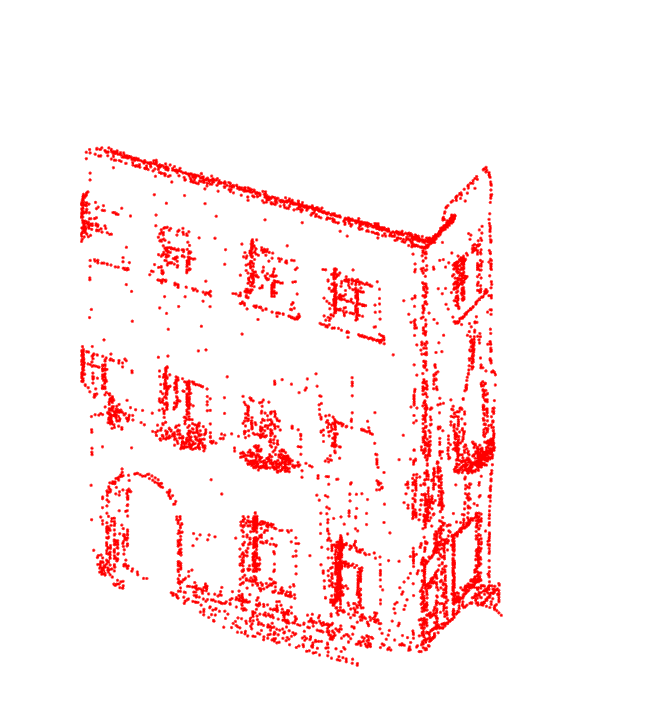}  
\\
{\small (a) Uniform resampling.} &  {\small (b) Contour-enhanced resampling.}
\\
\end{tabular}
  \end{center}
  \caption{\label{fig:crown}  Proposed resampling strategy enhances contours of a point cloud. Plots (a) and  (b) resamples $2\%$ points from a 3D point cloud of a building containing $381,903$ points. Plot (b) is more visual-friendly than Plot (a). Note that the proposed resampling strategy  is able to  to enhance any information depending on users' preferences.  }
\end{figure}

To solve this problem, an approach is to consider efficient data structures to represent 3D point clouds. For example,~\cite{RydeH:10, Loop:13} partitions the 3D space into voxels and discretizes point clouds over voxels; a drawback is that to achieve a fine resolution, a dense grid is required, which causes space inefficiency.~\cite{PengK:05,HornungWBSB:13} presents an octree representation of point clouds, which is space efficient, but suffers from discretization errors.~\cite{EckartK:16, EckartKTKK:16} presents a probabilistic generative model to model the distribution of point clouds; drawbacks are that those parametric models may not capture the true surface, and it is inefficient to infer parameters in the probabilistic generative model.

Another approach is to consider reducing the number of points through mesh simplification.  The main idea is to construct a triangular or polygonal mesh for 3D point clouds, where nodes are 3D points (need not be from the input points) and edges are connectivities between those points respecting certain restrictions (e.g., e.g. belonging to a manifold).  The mesh is simplified by reducing the number of nodes or edges; that is, several nodes are merged into one node with local structure preserved. Surveys of many such methods can be found in~\cite{PengKK:05, AlliezG:03, MagloLDH:15}. Drawbacks of this approach are that mesh construction requires costly computation, and mesh simplification changes the positions of original points, which causes distortion.

In this paper, we consider resampling 3D point clouds; that is, we design application-dependent resampling strategies to preserve application-dependent information. For example, conventional contour detection in 3D point clouds requires careful and costly computation to obtain surface normals and classification models~\cite{HackelWS:16, EckartKTKK:16}. We efficiently resample a small subset of points that is sensitive to the required contour information, making the subsequent processing cheaper without losing accuracy; see Figure~\ref{fig:crown} for an example. Since the original 3D point cloud is sampled from an object, we call this task~\emph{resampling}. This approach reduces the number of 3D points without changing the locations of original 3D points. After resampling, we unavoidably lose information in the original 3D point cloud.

The proposed method is rooted in rooted in graph signal processing, which is a framework to explore the interaction between signals and graph structure~\cite{SandryhailaM:13, ShumanNFOV:13}. We use a graph to capture local dependencies among points,  representing a discrete version of the surface of an original object. The advantage of using a graph is to capture both local and global structure of point clouds. Each of the 3D coordinates and other attributes associated with 3D points is a graph signal indexed by the nodes of the underlying graph. We thus formulate a resampling problem as  graph signal sampling.  However, graph sampling methods usually select samples in a deterministic fashion, which solves nonconvex optimization problems to obtain samples sequentially and requires costly computation~\cite{ChenVSK:15, AnisAO:15, MarquesSLR:15, TsitsveroBL:15}. To reduce the computational cost, we propose an efficient randomized resampling strategy to select a subset of points. The main idea is to generate subsamples according to a non-uniform resampling distribution, which is both fast and provably preserves application-dependent information in the original 3D point cloud.

We first propose a general feature-extraction based resampling framework. We use a general feature-extraction operator to represent application-dependent information. Based on this feature-extraction operator, we quantify the quality of resampling by using a simple, yet general reconstruction error, where we can derive the exact mean square error.  We obtain the optimal resampling distribution by optimizing the mean square error. The proposed optimal resampling distribution is guaranteed to be shift/rotation/scale-invariant.

We next specify a feature extraction operator to be a graph filter and study the specific optimal resampling distributions based on all-pass, low-pass and high-pass graph filtering. In each case, we derive an optimal resampling distribution and validate the performance on both simulated and real data.  We further combine all the proposed techniques into an efficient surface reconstruction system based on graph filter banks, which enables us to enhance features in a 3D point cloud. 

We finally apply the proposed methods on three applications: large-scale visualization, accurate registration and robust shape modeling.  In large-scale visualization, we use the proposed high-pass graph filtering based resampling strategy to highlight the contours of buildings and streets in a urban scene, which avoids saturation problems in visualization; in accurate registration, we use the proposed high-pass graph filtering based resampling strategy to extract the key points of a sofa, which makes the registration precise; in robust shape modeling, we use the proposed low-pass graph filtering based resampling strategy to reconstuct a surface, which makes the reconstruction robust to noise. The performances in those three applications validate the effectiveness and efficiency of the proposed resampling methods.

\mypar{Contributions} 
This paper considers a widely-used task from a novel theoretical perspective. As a preprocessing step, resampling a large-scale 3D point cloud uniformly is widely used in many tasks of large-scale 3D point cloud processing and many commercial softwares; however, people treat this step heuristically. This paper considers resampling 3D points from a theoretical signal processing perspective. For example, our theory shows that uniform resampling is the optimal resampling distribution when all 3D points are associated with the same feature values. The main contributions of the paper are as follows:
We propose
\begin{itemize}
\item a novel theoretical resampling framework for 3D point clouds with exact mean square error and optimal resampling distribution;
\item a novel feature-extraction operator for 3D point clouds based on graph filtering;
\item extensive empirical studies of the proposed resampling strategies on both simulated data and real point clouds.
\end{itemize}

This paper also points out many possible future directions of 3D point cloud processing, such as efficient 3D point cloud compression system based on graph filter banks, surface reconstruction based on arbitrary graphs and robust metric to evaluate the visualization quality of a 3D point cloud.

\mypar{Outline of the paper} Section~\ref{sec:formulation} formulates the resampling problem and briefly reviews
graph signal processing.
Section~\ref{sec:resample_feat} proposes a resampling framework based on general feature-extraction operator  and  Section~\ref{sec:resampe_filter} considers a graph filter as a specific feature-extraction operator. Three applications are presented in Section~\ref{sec:app}. Section~\ref{sec:conclusions} concludes the paper and provides pointers to future directions.

 \vspace{-2mm}
\section{Problem Formulation}
\label{sec:formulation}
In this section, we cover the background material necessary for the rest of the paper. We start with formulating a task of resampling a 3D point cloud. We then introduce graph signal processing, which lays a foundation for our proposed methods.

 \vspace{-4mm}
\subsection{Resampling a Point Cloud}
We consider a matrix representation of a point cloud with $N$ points and $K$ attributes,
\begin{eqnarray}
\label{eq:PC}
\X \ = \   \begin{bmatrix}
\s_1 & \s_2 & \ldots  &  \s_K
\end{bmatrix} \ = \ \begin{bmatrix}
\x_1^T \\ \x_2^T \\ \vdots  \\  \x_N^T
\end{bmatrix} \ \in \ \R^{N \times K},
\end{eqnarray}
where $\s_i \in \R^N$ represents the $i$th attribute and $\x_i \in \R^K$ represents the $i$th point. Depending on the sensing device, attributes can be 3D coordinates, RGB colors, textures, and many others. To distinguish 3D coordinates from other attributes, $\Xc \in \R^{N \times 3}$ represents 3D coordinates and $\Xo \in \R^{N \times (K-3)}$ represents other attributes.

The number of points $N$ is usually large. For example, a 3D scan of a building usually needs billions of 3D points. It is challenging to work with such a large-scale point cloud from both storage and data analysis perspectives. In many applications, however, we are interested in a subset of 3D points with particular properties, such as key points in point cloud registration and contour points in contour detection.  To reduce the storage and computational cost, we consider resampling a subset of representative 3D points from the original 3D point cloud to reduce the scale. The procedure of resampling is to resample $M~(M < N)$ points from a point cloud, or select $M$ rows from the point cloud matrix $\X$. The resampled point cloud is
$
\X_\M  \ = \ \Psi \X \ \in \R^{M \times K},
$
where $\M = (\M_1, \ldots , \M_{M})$ denotes the sequence of resampled indices,  called~\emph{resampled set}, $\M_i \in \{1, \ldots, N\}$ with $|\M| = M$ and the resampling operator $\Psi$ is a linear mapping from
$\R^N$ to $\R^M$, defined as
\begin{equation}
\label{eq:Psi}
 \Psi_{i,j} = 
  \left\{ 
    \begin{array}{rl}
      1, & j = \M_i;\\
      0, & \mbox{otherwise}.
  \end{array} \right. 
\end{equation}

The efficiency of the proposed resampling strategy is critical. Since we work with a large-scale point cloud, we want to avoid expensive computation. To implement resampling in an efficient way, we consider a randomized resampling strategy. It means that the resampled indices are chosen according to a resampling distribution. Let $\{\pi_{i}\}_{i=1}^N$ be a series of resampling probabilities, where $\pi_i$ denotes the probability to select the $i$th sample in each random trial. Once the resampling distribution is chosen, it is efficient to generate samples. The goal here is to find a resampling distribution that preserves information in the original point cloud.

The invariant property of the proposed resampling strategy is also critical. When we shift, rotate or scale a point cloud, the intrinsic distribution of 3D points does not changed and the proposed  resampling strategy should not change.

\begin{defn}
\label{def:shift}
A resampling strategy is shift-invariant when a sampling distribution $\pi$ is designed for a point cloud, $\X = \begin{bmatrix}
\Xc & \Xo
\end{bmatrix}$, 
then the same sampling distribution $\pi$ is designed for its shifted point cloud, 
$\begin{bmatrix}
\Xc + \one \a^T & \Xo
\end{bmatrix}
$ with $\a \in \R^3$.
\end{defn}

\begin{defn}
\label{def:rotation}
A resampling strategy is rotation-invariant when a sampling distribution $\pi$ is designed for a point cloud, $\X = \begin{bmatrix}
\Xc & \Xo
\end{bmatrix}$, then the same sampling distribution  $\pi$ is designed for its rotated point cloud, 
$\begin{bmatrix}
\Xc \RR & \Xo
\end{bmatrix}
$, where $\RR \in \R^{3 \times 3}$ is a 3D rotation matrix.
\end{defn}

\begin{defn}
\label{def:scale}
A resampling strategy is scale-invariant when a sampling distribution $\pi$ is designed for a point cloud, $\X = \begin{bmatrix}
\Xc & \Xo
\end{bmatrix}$, then the same sampling distribution  $\pi$ is designed for its rotated point cloud, 
$\begin{bmatrix}
c \Xc & \Xo
\end{bmatrix}
$, where constant $c>0$.
\end{defn}
Our aim is to guarantee that the proposed resampling strategy is shift, rotation and scale  invariant.

\subsection{Graph Signal Processing for Point Clouds}
A graph is a natural and efficient way to represent a 3D point cloud because it represents a discretized version of an original surface. In computer graphics, polygon meshes, as a class of graphs with particular connectivity restrictions,  are extensively used to represent the shape of an object~\cite{FlorianiM:02}; however, mesh construction usually requires sophisticated geometry analysis, such as calculating surface normals, and the mesh representation may not be the most suitable representation for analyzing point clouds because of connectivity restrictions. Here we extend polygon meshes to general graphs by relaxing the connectivity restrictions. Such graphs are easier to construct and are flexible to capture geometry information.

\mypar{Graph Construction}
We construct a general graph of a point cloud by encoding the local geometry information through an adjacency matrix $\W \in \R^{N \times N}$.  Let $\xc_i \in \R^3$ be the 3D coordinates of the $i$th point; that is, the $i$th row of $\Xc$. The edge weight between two points $\xc_i$ and $\xc_j$ is 
\begin{equation}
\label{eq:adj}
\W_{i,j} = 
\left\{ 
    \begin{array}{rl}
      e^{- \frac{\left\| \xc_i - \xc_j \right\|_2^2}{\sigma^2} }, & \left\| \xc_i - \xc_j \right\|_2 \leq \tau;\\
      0, & \mbox{otherwise},
  \end{array} \right. 
\end{equation}
where variance $\sigma$ and threshold $\tau$ are parameters. Equation~\eqref{eq:adj} shows that when the Euclidean distance of two points is smaller than a threshold $\tau$, we connect these two points by an edge and the edge weight depends on the similarity of two points in the 3D space. The weighted degree matrix $\D$ is a diagonal matrix with diagonal element $\D_{i,i} = \sum_{j} \W_{i,j}$ reflecting the density around the $i$th point. This graph is approximately a discrete representation of the original surface and can be efficiently constructed via a tree data structure, such as octree~\cite{PengK:05,HornungWBSB:13}. Here we only use the 3D coordinates to construct a graph, but it is also feasible to take other attributes into account~\eqref{eq:adj}. Given this graph, the attributes of point clouds are called~\emph{graph signals}. For example, an attribute $\s$ in~\eqref{eq:PC} is a signal index by the graph.

\mypar{Graph Filtering}  A graph filter is a system that takes a graph signal as an input and produces another graph signal as an output. Let  $\Adj \in \R^{N \times N}$ be a~\emph{graph shift operator}, which is the most elementary nontrivial graph filter. Some common choice of a graph shift operator  is the adjacency matrix $\W$~\eqref{eq:adj}, the transition matrix $\D^{-1} \W$, the graph Laplacian matrix $\D-\W$, and many other structure-related matrices. The graph shift replaces the signal value at a node with a weighted linear combination
of values at its neighbors; that is,
$
\y \ = \ \Adj \s \ \in \ \R^N,
$
where $\s \in \R^N$ is an input graph signal (an attribute of a point cloud). Every linear, shift-invariant graph filter is a polynomial in the graph shift~\cite{SandryhailaM:13}
\begin{equation}
\label{eq:graph_filter}
h(\Adj) \ = \ \sum_{\ell = 0} ^{L-1}  h_{\ell} \Adj^{\ell} = h_0\Id + h_1\Adj + \ldots + h_{L-1}\Adj^{L-1},
\end{equation}
where $h_{\ell} (\ell = 0, 1, \ldots, L-1)$ are filter coefficients and $L$ is the length of this graph filter. Its output is given by the matrix-vector product
$
\y \ = \ h(\Adj) \s \ \in \ \R^N.
$

\mypar{Graph Fourier Transform}  The eigendecomposition of a graph shift operator $\Adj$ is~\cite{SandryhailaM:131}
\begin{equation}
  \label{eq:eigendecomposition}
  \Adj \ = \ \Vm \Lambda \Vm^{-1},
\end{equation}
where the  eigenvectors of $\Adj$ form the columns of matrix $\Vm$,  and the eigenvalue matrix
$\Lambda \in \R^{N\times N}$ is the diagonal matrix of corresponding
eigenvalues $\lambda_1, \, \ldots, \, \lambda_{N}$ of $\Adj$ ($\lambda_1 \geq  \lambda_2 \geq \ldots, \,  \geq \lambda_{N} )$.  These eigenvalues represent frequencies on the graph~\cite{SandryhailaM:131} where $\lambda_1$ is the lowest frequency and $\lambda_{N}$ is the highest frequency.  Correspondingly, $\vv_1$ captures the smallest variation on the graph and $\vv_N$ captures the highest variation on the graph. $\Vm$ is also called~\emph{graph Fourier basis}. The~\emph{graph Fourier transform} of a graph signal $\s \in \R^N$ is
\begin{equation}
  \label{eq:graph_FT}
  \widehat{\s} \ = \ \Vm^{-1} \s.
\end{equation}
 The~\emph{inverse graph Fourier transform} is
$
 \s  \ = \  \Vm  \widehat{\s}  = \sum_{k=1}^{N}  \widehat{s}_k \vv_k,
$
where $\vv_k$ is the $k$th column of $\Vm$ and $\widehat{s}_k $ is the $k$th component in $\widehat{\s}$. The vector $\widehat{\s}$ in~\eqref{eq:graph_FT} represents the
signal's expansion in the eigenvector basis and describes the frequency
components of the graph signal $\s$. The inverse graph Fourier transform
reconstructs the graph signal by combining
graph frequency components.

\section{Resampling based on Feature Extraction }
\label{sec:resample_feat}

During resampling, we reduce the number of points and unavoidably lose information in a point cloud.  Our goal is to design an application-dependent resampling strategy, preserving selected information depending on particular needs. Those information are described by features. When detecting contours, we usually need careful and intensive computation, such as calculating surface normals and classifying points~\cite{HackelWS:16, EckartKTKK:16}. Instead of working with a large number of points, we consider efficiently sampling a small subset of points that captures the required contour information, making the subsequent computation much cheaper without losing contour information. We also need to guarantee that the proposed resampling strategy is shift/rotation/scale-invariant for robustness. We will show that some features naturally provide invariance and other may not. We will handle the invariance by considering a general objective function.

\subsection{Feature-Extraction based Formulation}
Let $f(\cdot) $ be a feature-extraction operator that extracts targeted information from a point cloud according to particular needs; that is, the features $f(\X) \in \R^{N \times K}$ are extracted from a point cloud $\X \in \R^{N \times K}$\footnote{
For simplicity, we consider the number of features to be the same as the number of attributes. The proposed method also works when the number of features and the number of attributes are different.}. Depending on an application, those features can be edges, key points and flatness~\cite{WeberHH:10,  DanielsHOS:07, ChoiTC:13, TaguchiJRF:13, FengTK:14}. In this section, we consider feature-extraction operator at an abstract level and use graph filters to implement a feature-extraction operator  in the next section.

To evaluate the performance of a resampling operator, we quantify how much features are lost during resampling; that is, we sample features, and then interpolate to get back original features. The features are considered to reflect the targeted information contained in each 3D point. The performance is better when the recovery error is smaller. Mathematically,  we resample a point cloud $M$ times. At the $j$th step, we independently choose a point $\M_j = i$ with probability $\pi_{i}$. Let $\Psi \in \R^{M \times N}$ be the resampling operator~\eqref{eq:Psi} and $\Ss \in \R^{N \times N}$ be a diagonal rescaling matrix with $\Ss_{i,i} = 1/\sqrt{ M \pi_i }$.  We 
quantify the performance of a resampling operator as follows:
\begin{equation}
\label{eq:att_obj}
 D_{f(\X)} (\Psi) \ = \ \left\| \Ss \Psi^T \Psi f(\X)  - f(\X)  \right\|_2^2,
\end{equation}
where $ \left\| \cdot \right\|_2$ is the spectral norm.  $\Psi^T \Psi \in \R^{N \times N}$ is a zero-padding operator, which a diagonal matrix with diagonal elements $(\Psi^T \Psi)_{i,i} > 0$ when the $i$th point is sampled, and $0$, otherwise. The zero-padding operator $\Psi^T \Psi $ ensures the resampled points and the original point cloud have the same size.  $\Ss$ is used to compensate non-uniform weights during resampling.  $\Ss \Psi^T $ is the most naive interpolation operator that reconstructs  the original feature $f(\X)$ from its resampled version $\Psi f(\X) $ and $\Ss \Psi^T \Psi f(\X) $ represents the preserved features after resampling in a zero-padding form. Lemma~\ref{lem:unbias} shows that $\Ss$ aids to provide an unbiased estimator. 
\begin{myLem} 
\label{lem:unbias}
Let $f(\X) \in \R^{N \times K}$ be features extracted from a point cloud $\X$. Then,
\begin{eqnarray*}
\mathbb{E}_{\Psi \sim \pi} \left(  \Psi^T \Psi f(\X) \right) & \propto & \pi \odot f(\X),
\\
	\mathbb{E}_{\Psi \sim \pi}  \left(  \Ss \Psi^T \Psi f(\X) \right)  & = &  f(\X),
\end{eqnarray*}
where $\mathbb{E}_{\Psi \sim \pi}$ means the expectation over samples, which are generated from a distribution $\Pi$ independently and randomly, and $\odot$ is row-wise multiplication. 
\end{myLem}
The proof is shown in Appendix~\ref{sec:app1}.

The evaluation metric $D_{f(\X)} (\Psi)$ measures the reconstruction error; that is, how much feature information is lost after resampling without using sophisticated interpolation operator. When $D_{f(\X)} (\Psi)$ is small, preserved features after resampling are close to the original features, meaning that little information is lost. The expectation $\mathbb{E}_{\Psi \sim \pi} \left( D_{f(\X)} (\Psi) \right)$ is the expected error caused by resampling and quantifies the performance of a resampling distribution $\pi$. Our goal is to minimize $\mathbb{E}_{\Psi \sim \pi} \left(  D_{f(\X)} (\Psi) \right)$ over $\pi$ to obtain an optimal resampling distribution in terms of preserving features $f(\X)$. We now derive the mean square error of the objective function~\eqref{eq:att_obj}.

\begin{myThm} 
\label{thm:MSE}
The mean square error of the objective function~\eqref{eq:att_obj} is
\begin{equation}
\label{eq:MSE}
\mathbb{E}_{\Psi \sim \pi}  D_{f(\X)} (\Psi) \ = \
 {\rm Tr} \left(  f(\X) \Q  f(\X)^T   \right),
\end{equation}
where $\Q \in \R^{N \times N}$  is a diagonal matrix with $\Q_{i,i} = 1/\pi_i - 1$.
\end{myThm}
The proof is shown in Appendix~\ref{sec:app2}.

We now consider the invariance property of resampling. The sufficient condition for the shift/rotation/scale-invariance of a resampling strategy is that  the evaluation metric~\eqref{eq:att_obj} be  shift/ rotation/scale-invariance. Recall that  a 3D point cloud is $\X = \begin{bmatrix}
\Xc & \Xo
\end{bmatrix}$,
where $\Xc \in \R^{N \times 3}$ represents 3D coordinates and $\Xo \in \R^{N \times (K-3)}$ represents other attributes.

\begin{defn}
\label{def:shift_feature}
A feature-extraction operator $f(\cdot)$ is shift-invariant when the features extracted from a point cloud and its shifted version are same; that is, $f(\begin{bmatrix}
\Xc & \Xo
\end{bmatrix}) = f(\begin{bmatrix}
\Xc + \one \a^T & \Xo
\end{bmatrix})$ with shift $\a \in \R^3$.
\end{defn}

\begin{defn}
\label{def:rotation_feature}
A feature-extraction operator $f(\cdot)$ is rotation-invariant when the features extracted from a point cloud and its rotated version are same; that is, $f(\begin{bmatrix}
\Xc & \Xo
\end{bmatrix}) = f(\begin{bmatrix}
\Xc \RR & \Xo
\end{bmatrix})$ with $\RR \in \R^{3 \times 3}$ is a 3D rotation matrix.
\end{defn}

\begin{defn}
\label{def:scale_feature}
A feature-extraction operator $f(\cdot)$  is scale-invariant when features extracted from a point cloud and its scaled version are same; that is,
$f(\begin{bmatrix}
\Xc & \Xo
\end{bmatrix}) = f(\begin{bmatrix}
c \Xc & \Xo
\end{bmatrix})$ with constant $c >0$.
\end{defn}

When $f(\cdot)$ is shift/rotation/scale-invariant,~\eqref{eq:att_obj}  does not change through shifting, rotating or scaling, leading to a shift/rotation/scale-invariant  resampling strategy and it is sufficient to minimize $\mathbb{E}_{\Psi \sim \pi} \left( D_{f(\X)} (\Psi) \right)$ to obtain a resampling strategy; however, when $f(\cdot)$ is shift/rotation/scale-variance,~\eqref{eq:att_obj}  may change through shifting, rotating or scaling, leading to a shift/rotation/scale-variant resampling strategy.

To handle shift variance, we can  recenter a point cloud to the origin before processing; that is, we normalize the mean of 3D coordinates to zeros.  To handle scale variance, we can  normalize  
the magnitude of the 3D coordinates before processing; that is, we normalize the spectral norm $\left\|  \X_c \right\|_2 = c$ with constant $c>0$. The choice of $c$ depends on users' preference and we will show that $c$ is a trade-off between 3D coordinates and the values of other attributes. From now on, we first recenter a point cloud to the origin and then normalize its magnitude to guarantee the shift/scale invariance of any 3D point cloud.

To handle rotation variance of $f(\cdot)$, we consider the following evaluation metric:
\begin{eqnarray}
\label{eq:variant_att_obj}
 D_f(\Psi) & = & \max_{\X'_c: \left\|  \X'_c \right\|_2 = c }  D_{f \left( \begin{bmatrix}
  \X'_c & \Xo
\end{bmatrix} \right) } 
\left( \Psi \right)
\nonumber  \\
& = & \max_{\X'_c: \left\|  \X'_c \right\|_2 = c }    \left\| 
  \left( \Ss \Psi^T \Psi - \Id \right)  f \left(
  \begin{bmatrix}
  \X'_c & \Xo
\end{bmatrix} \right) \right\|_F^2,
\nonumber
\\
\end{eqnarray}
where constant $c = \left\|  \X_c \right\|_2$  is the normalized spectral norm of 3D coordinates. 

Unlike $ D_{f(\X)}(\Psi)$~\eqref{eq:att_obj},  to remove the influence of rotation, the evaluation metric $ D_f(\Psi)$ considers the  worst possible reconstruction error caused by rotation.  In~\eqref{eq:variant_att_obj}, we consider 3D coordinates as variables due to rotation. We constrain the spectral norm of 3D coordinates because a rotation matrix is orthornormal and the spectral norm of 3D coordinates does not change during rotation.  We then minimize $\mathbb{E}_{\Psi \sim \pi} \left( D_f (\Psi) \right)$ to obtain a rotation-invariant resampling strategy even when $f(\cdot)$ is rotation-variant.

For simplicity, we perform derivation for only linear feature-extraction operators.  A linear feature-extraction operator $f(\cdot)$ is of the form of  $f(\X) = \F \X$, where $\X$ is a 3D point cloud and $\F \in \R^{N \times N}$ is a feature-extraction matrix.

\begin{myThm}
\label{thm:MSE_variant}
Let $f(\cdot) $ be a rotation-varying linear feature-extraction operator, where $f(\X) = \F \X$ with $\F \in \R^{N \times N}$.  The exact form of $\mathbb{E}_{\Psi \sim \pi}  D_{f} (\Psi)$ is
\begin{equation}
\label{eq:MSE_variant}
\mathbb{E}_{\Psi \sim \pi} \left(  D_{f} (\Psi) \right) \ = \
 c^2 {\rm Tr} \left( \F \Q  \F^T \right) +  {\rm Tr} \left( \F \Xo \Q  (\F\Xo)^T \right),
\end{equation}
where $\Q \in \R^{N \times N}$  is a diagonal matrix with $\Q_{i,i} = 1/\pi_i - 1$.
\end{myThm}
The proof is shown in Appendix~\ref{sec:app3}.

\subsection{Optimal Resampling Distribution}
We now derive the optimal resampling distributions by minimizing the reconstruction error. For a rotation-invariant feature-extraction operator, we minimize~\eqref{eq:MSE}.
\begin{myThm}
\label{thm:opt_sample_invariant}
Let $f(\cdot)$ be a rotation-invariant feature-extraction operator. The corresponding optimal resampling strategy $\pi^*$ is, 
\begin{equation}
\label{eq:opt_sample_inv}
 \pi^*_i  \propto \left\| f_i (\X) \right\|_2,
\end{equation}
where $f_i (\X) \in \R^K$ is the $i$th row of $f (\X)$.
\end{myThm}
The proof is shown in Appendix~\ref{sec:app4}. We see that the optimal resampling distribution is proportional to the magnitude of features; that is, points associated with high magnitudes have high probability to be selected, while points associated with small magnitudes have small probability to be selected. The intuition is that the response after the feature-exaction operator reflects the information contained in each 3D point and determines the resampling probability of each 3D point.

For a rotation-variant linear feature-extraction operator, we minimize~\eqref{eq:MSE_variant}.
\begin{myThm}
\label{thm:opt_sample_variant}
Let $f(\cdot) $ be a rotation-variant linear feature-extraction operator, where $f(\X) = \F \X$ with $\F \in \R^{N \times N}$. The corresponding optimal resampling strategy $\pi^*$ is, 
\begin{equation}
\label{eq:opt_sample_variant}
 \pi^*_i  \propto   \sqrt{  c^2 \left\| \F_i \right\|_2^2  +   \left\| \left( \F \Xo \right)_i \right\|_2^2 },
\end{equation}
where constant $c = \left\| \Xc \right\|_2$, $\F_i$ is the $i$th row of $\F$ and $\left( \F \Xo \right)_i $ is the $i$th row of $\F \Xo $.
\end{myThm}
The proof is shown in Appendix~\ref{sec:app5}.  We see that the optimal resampling distribution is also proportional to the magnitude of features. The feature comes from two sources: 3D coordinates and the other attributes. The tuning parameter $c$ in~\eqref{eq:opt_sample_variant} is the normalized spectral norm used to remove the scale variance. The choice of $c$ trade-offs the contribution from  3D coordinates and the other attributes.

\section{Resampling based on Graph Filtering}
\label{sec:resampe_filter}

The previous section studied resampling based on an arbitrary feature-extraction operator. In this section, we design graph filters to efficiently extract features from a point cloud.  Let features extracted from a point cloud $\X$ be
\begin{equation*}
f(\X) \ = \ h(\Adj) \X \ = \  \sum_{\ell = 0} ^{L-1}  h_{\ell} \Adj^{\ell}  \X,
\end{equation*}
which follows from the definition of graph filters~\eqref{eq:graph_filter}.  Since a graph filter is a linear operator, the corresponding optimal resampling distribution follows from the results in Theorems~\ref{thm:opt_sample_invariant} and~\ref{thm:opt_sample_variant} by replacing $\F = \sum_{\ell = 0} ^{L-1}  h_{\ell} \Adj^{\ell}$. All graph filtering-based feature-extraction operators are scale-variant due to linearity. As discussed earlier, we can normalize the spectral norm of a 3D coordinates to handle this issue. We thus will not discuss scale invariance in this section. We will see that by carefully using the graph shift operator $\Adj$ and filter coefficients $h_i$s, a graph filtering-based feature-extraction operator may be shift or rotation varying.

 Similarly to filter design in classical signal processing,  we design a graph filter either in the graph vertex domain or in the graph spectral domain. In the graph vertex domain, for each point, a graph filter averages the attributes of its local points.  For example, the output of the $i$th point, $f_i(\X) = \sum_{\ell = 0} ^{L-1}  h_{\ell} \left( \Adj^{\ell} \X \right)_i$ is a weighted average of the attributes of points that are within $L$ hops away from the $i$th point. The $\ell$th graph filter coefficient, $h_{\ell}$, quantifies the contribution from the $\ell$th-hop neighbors. We design the filter coefficients to change the weights in local averaging.

In the graph spectral domain, we first design a graph spectrum distribution and then use graph filter coefficients to fit this distribution. For example, a graph filter with length $L$ is
\begin{eqnarray*}
&& h(\Adj) \ = \ \Vm h(\Lambda) \Vm^{-1}
\\
& = &  \Vm 
\begin{bmatrix}
\sum_{\ell = 0} ^{L-1}  h_{\ell} \lambda_1^{\ell} &  0 & \cdots & 0  \\
0  &  \sum_{\ell = 0} ^{L-1}  h_{\ell} \lambda_2^{\ell} & \cdots & 0 \\
\vdots &  \vdots & \ddots & \vdots  \\
0  &  0 & \cdots & \sum_{\ell = 0} ^{L-1}  h_{\ell} \lambda_N^{\ell}
\end{bmatrix}
 \Vm^{-1},
\end{eqnarray*}
where $\Vm$ is the graph Fourier basis and $\lambda_i$ are graph frequencies~\eqref{eq:eigendecomposition}. When we want the response of the $i$th graph frequency to be $c_i$, we set 
$$
h(\lambda_i) \ = \ \sum_{\ell = 0} ^{L-1}  h_{\ell} \lambda_i^{\ell}
\ = \ c_i,
$$ and solve a set of linear equations to obtain the graph filter coefficients $h_{\ell}$. It is also possible to use the Chebyshev polynomial to design graph filter coefficients~\cite{HammondVG:11}.  We now consider some special cases of graph filters. 

\subsection{All-pass Graph Filtering} 
Let $h(\lambda_i) = 1$; that is, $h(\Adj) = \Id$ is an identity matrix with $h_0 = 1$ and $h_i = 0$ for $i = 1, \ldots, L-1$.  The intuition behind this setting is that the original point cloud is trustworthy and all points are uniformly sampled from an object without noise, reflecting the true geometric structure of the object. We  want to preserve all the information and the features are thus the original attributes themselves. Since $f(\X) = \X$, the feature-extraction operator $f(\cdot)$ is rotation-variant.  Based on Theorem~\ref{thm:opt_sample_variant}, the optimal resampling strategy is
\begin{equation}
 \pi^*_i  \propto   \sqrt{  c^2  +   \left\| \left( \Xo \right)_i \right\|_2^2 }.
\end{equation}
Here the feature-extraction matrix $\F$ in~\eqref{eq:opt_sample_inv} is an identity matrix and the norm of each row of $\F$ is 1.  When we only preserve 3D coordinates, we ignore the term of $\Xo$ and obtain a constant resampling probability for each point, meaning that uniform resampling is the optimal resampling strategy to preserve the overall geometry information.

\subsection{High-pass Graph Filtering}
In image processing,  a high-pass filter is used to extract edges and contours. Similarly, we use a high-pass graph filter to extract contours in a point cloud.  Here we only consider the 3D coordinates as attributes ($\X = \Xc = \R^{N \times 3}$), but the proposed method can be easily extended to other attributes.

A critical question is how to define contours in a 3D point cloud. We consider that contour points break the trend formed by its neighboring points and bring innovation. Many previous works need sophisticated geometry-related computation, such as surface normal, to detect contours~\cite{HackelWS:16}. Instead of measuring sophisticated geometry properties,  we describe the possibility of being a contour point by the local variation on graphs, which is the response of high-pass graph filtering. The corresponding local variation of the $i$th point is 
\begin{equation}
\label{eq:LV}
f_i (\X)  = \left\| \ \left( h(\Adj) \X \right)_i \right\|_2^2,
\end{equation}
where $h(\Adj)$ is a high-pass graph filter. The local variation $f(\X) \in \R^N$ quantifies the energy of response after high-pass graph filtering. The intuition behind this is that when the local variation of a point is high, its 3D coordinates cannot be well approximated from the 3D coordinates of its neighboring points; in other words, this point bring innovation by breaking the trend formed by its neighboring points and has a high possibility of being a contour point.

The following theorem shows  that in general the local variation is rotation invariant, but shift variant.
\begin{myThm}
\label{thm:feature_invariant}
Let $f(\X) = {\rm diag} \left( h (\Adj)  \X \X^T h (\Adj)^T \right) \in \R^N$, where diag$(\cdot)$ extracts the diagonal elements. $f(\X)$ is rotation invariant and shift invariant unless $h(\Adj) \one = \bold{0} \in \R^N$.
\end{myThm}
The proof is shown in Appendix~\ref{sec:app6}.

To guarantee that local variation is naturally shift invariant without recentering a 3D point cloud,  we simply use a transition matrix as a graph shift operator; that is, $\Adj = \D^{-1} \W$, where $\D$ is the diagonal degree matrix. The reason is that $\one \in \R^N$ is the eigenvector of a transition matrix, $\Adj \one = \D^{-1} \W \one = \one$. Thus, 
\begin{eqnarray*} 
h(\Adj) \one = \sum_{\ell = 0} ^{N-1}  h_{\ell} \Adj^{\ell} \one =  \sum_{\ell = 0} ^{N-1}  h_{\ell}  \one  = \bold{0},
\end{eqnarray*}
when $\sum_{\ell = 0} ^{N-1}  h_{\ell}  = 0$. A simple design is a Haar-like high-pass graph filter
\begin{eqnarray}
\label{eq:HH}
 h_{\rm HH}(\Adj) 
& = & \Id - \Adj
\\
\nonumber
& = &  \Vm 
\begin{bmatrix}
1 - \lambda_1 &  0 & \cdots & 0  \\
0  &  1 -  \lambda_2 & \cdots & 0 \\
\vdots &  \vdots & \ddots & \vdots  \\
0  &  0 & \cdots & 1 -  \lambda_N
\end{bmatrix}
 \Vm^{-1},
\end{eqnarray}
Note that $\lambda_{\rm max} = \max_{i} |\lambda_i| = 1$, where $\lambda_i$ are eigenvalues of $\Adj$,  because the graph shift operator is a transition matrix. In this case, $h_0 = 1, h_1 = -1$ and $h_i = 0$ for all $i>1$, $\sum_{\ell = 0} ^{N-1}  h_{\ell}  = 0$. Thus, a Haar-like high-pass graph filter is both shift and rotation invariant.   The graph frequency response of a Haar-like high-pass graph filter is $h_{\rm HH}(\lambda_i) = 1- \lambda_i$. Since the eigenvalues are ordered descendingly, we have $1- \lambda_i \leq 1- \lambda_{i+1}$, meaning low frequency response attenuates and high frequency response amplifies.

\begin{figure}[htb]
  \begin{center}
    \begin{tabular}{cc}
\includegraphics[width=0.4\columnwidth]{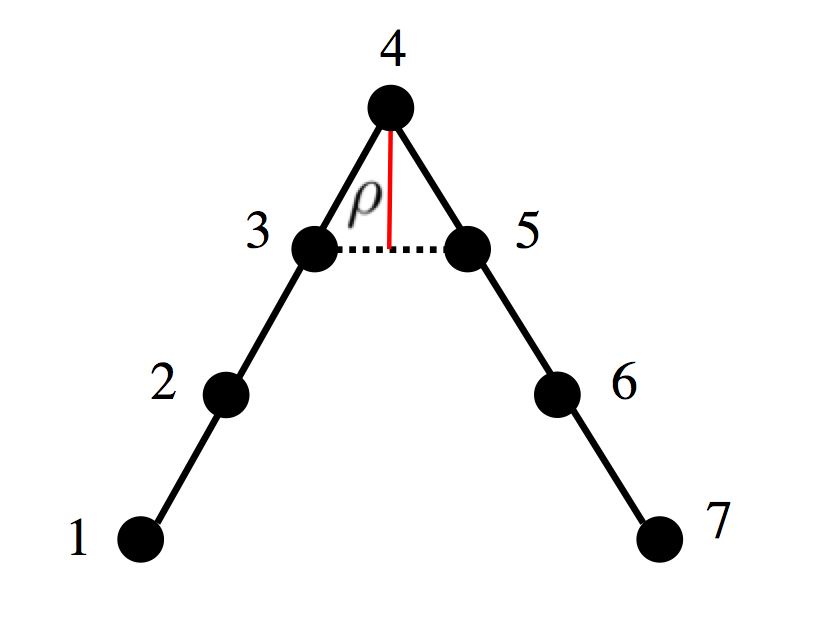}  & \includegraphics[width=0.36\columnwidth]{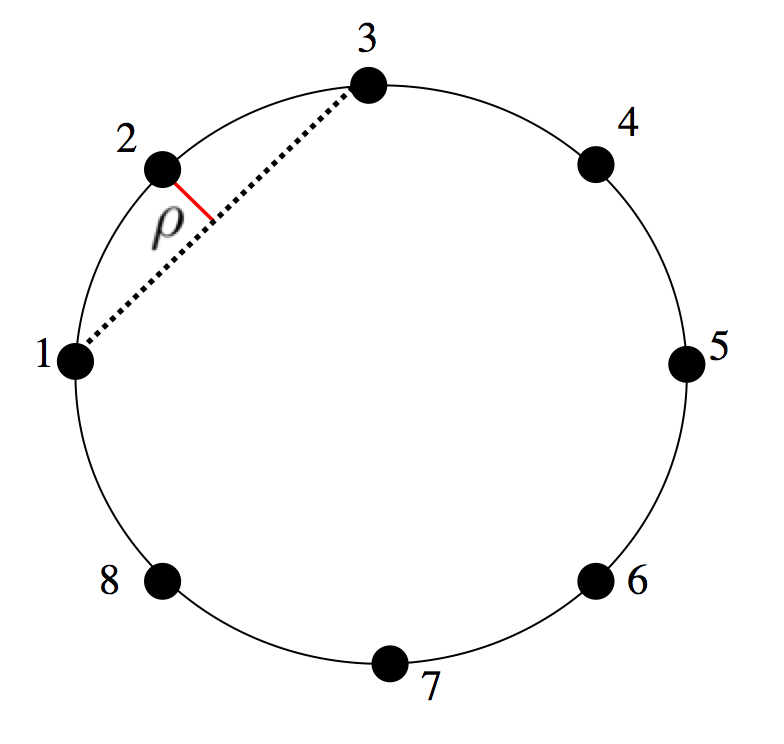} 
\\
      {\small (a)  Lines. } & {\small (b) Circle. } 
\end{tabular}
  \end{center}
  \caption{\label{fig:innovation}  Red line shows the local variation. }
    \vspace{-3mm}
\end{figure}

In the graph vertex domain, the response of the $i$th point is
\begin{equation*}
\left( h_{\rm HH}(\Adj)  \X \right)_i \ = \   \x_i - \sum_{j \in \N_i} \Adj_{i,j} \x_j.
\end{equation*}
Because $\Adj$ is a transition matrix,  $\sum_{j \in \N_i} \Adj_{i,j} = 1$ and  $h_{\rm HH}(\Adj)$ compares the difference between a point and the convex combination of its neighbors.  The geometry interpretation of the proposed local variation is the Euclidean distance between the original point and the convex combination of its neighbors, reflecting how much  information we know about a point from its neighbors. When the local variation of a point is large, the Euclidean distance between this point and the convex combination of its neighbors is long and this point provides a large amount of variation.

We can verify the proposed local variation on some simple examples.

\begin{axiom}
\label{pri:line}
When a point cloud forms a 3D line, two endpoints belong to the contour.
\end{axiom}
\begin{axiom}
\label{pri:polygon}
When a point cloud forms a 3D polygon/polyhedron, the vertices (corner points) and the edges (line segment connecting two adjacent vertices)  belong to the contour.
\end{axiom}
\begin{axiom}
\label{pri:circle}
When a point cloud forms a 3D circle/sphere, there is no contour.
\end{axiom}

When the points are uniformly spread along the defined shape, the proposed local variation~\eqref{eq:LV} satisfies Examples~\ref{pri:line},~\ref{pri:polygon} and~\ref{pri:circle} from the geometric perspective.  In Figure~\ref{fig:innovation} (a), Point 2 is the convex combination of Points 1 and 3, and the local variation of Point 2 is thus zero. However, Point 4 is not the convex combination of Points 3 and 5 and the length of the red line indicates the local variation of Point 4. Only Points 1, 4 and 7 have nonzero local variation, which is what we expect. In Figure~\ref{fig:innovation} (b), all the nodes are evenly spread on a circle and have the same amount of variation, which is represented as a red line. Similar arguments show that the proposed local variation~\eqref{eq:LV} satisfies Examples~\ref{pri:line},~\ref{pri:polygon} and~\ref{pri:circle}.

The feature-extraction operator $f(\X) = \left\| h_{\rm HH} (\Adj) \X \right\|_F^2$ is shift and rotation-invariant. Based on Theorem~\ref{thm:opt_sample_invariant}, the optimal resampling distribution is 
\begin{eqnarray}
\label{eq:opt_sample_HH}
 \pi^*_i  & \propto  &  \bigg\| \left( h_{\rm HH} (\Adj) \X \right)_i \bigg\|_2^2 \ = \  \left\| \x_i - \sum_{j \in \N_i} \Adj_{i,j} \x_j  \right\|_2^2,
\end{eqnarray}
where $\Adj = \D^{-1} \W$ is a transition matrix.

\begin{figure}[htb]
  \begin{center}
\includegraphics[width=0.5\columnwidth]{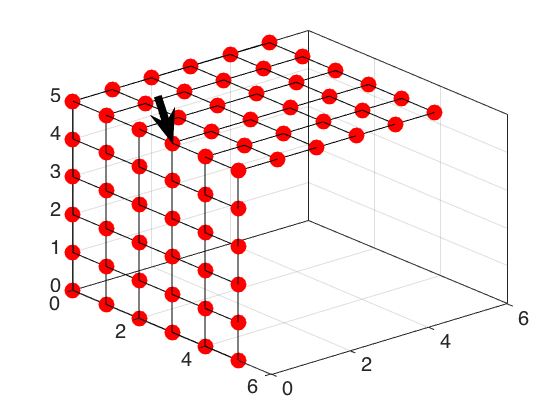} 
  \end{center}
  \caption{\label{fig:cube}  The pairwise difference based local variation cannot capture the contour points connecting two faces.  }
  \vspace{-3mm}
\end{figure}

Note that the graph Laplacian matrix is commonly used to measure  variations. Let $\LL = \Dd - \W \in \R^{N \times N}$ be a graph Laplacian matrix. The graph Laplacian based total variation is
\begin{equation}
 {\rm Tr} \left( \X^T \LL \X \right) \ = \  \sum_{i} \sum_{j \in \N_i } \W_{i,j} \left\| \x_i - \x_j \right\|_2^2.
\end{equation}
where $\N_i$ is the neighbors of the $i$th node and the variation contributed by the $i$th point is 
\begin{equation}
\label{eq:pair_LV}
f_i (\X) \ = \ \sum_{j \in \N_i } \W_{i,j} \left\| \x_i - \x_j \right\|_2^2.
\end{equation}
The  variation here is defined based on the accumulation of  pairwise differences. We call~\eqref{eq:pair_LV} pairwise difference based local variation. The pairwise difference based local variation cannot capture geometry change and violates Example~\ref{pri:polygon}.  We show a counter example in Figure~\ref{fig:cube}. The points are uniformly spread along the faces of a cube and Figure~\ref{fig:cube} shows two faces. Each point connects to its adjacent four points with the same edge weight. The pairwise difference based local variations of all the points are the same, which means that there is no contour in this point cloud. However, the black arrow points to a point that should be a contour point.

\begin{figure}[htb]
  \begin{center}
    \begin{tabular}{ll}
\includegraphics[width=0.42\columnwidth]{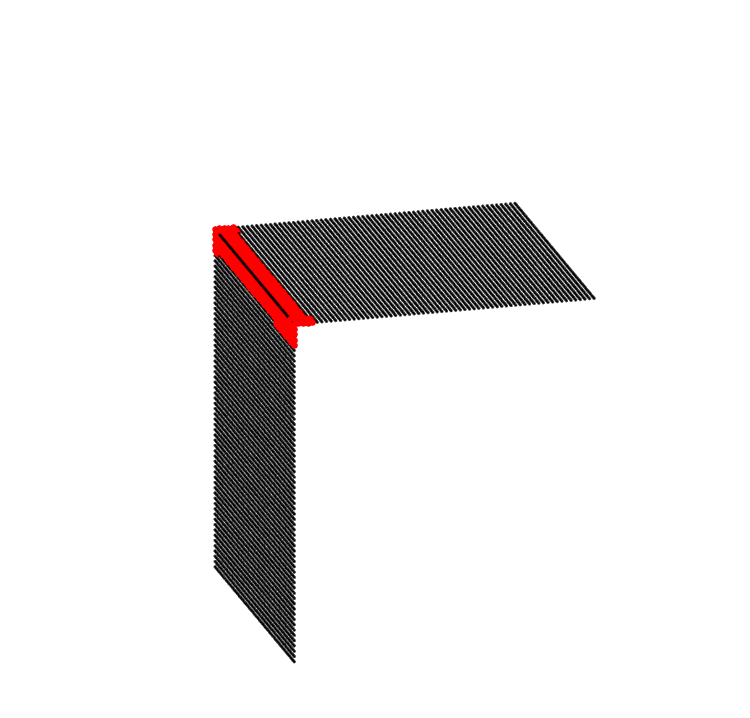}  
&
\includegraphics[width=0.42\columnwidth]{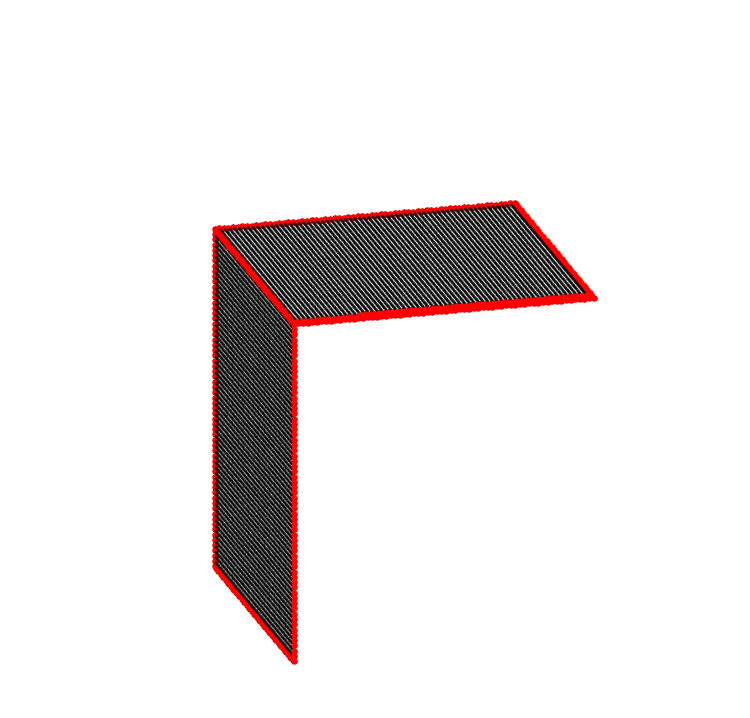}
\\
{\small (a) Hinge: Difference of normals. }  & {\small (b)  Hinge: Local variation. }
\\
\includegraphics[width=0.45\columnwidth]{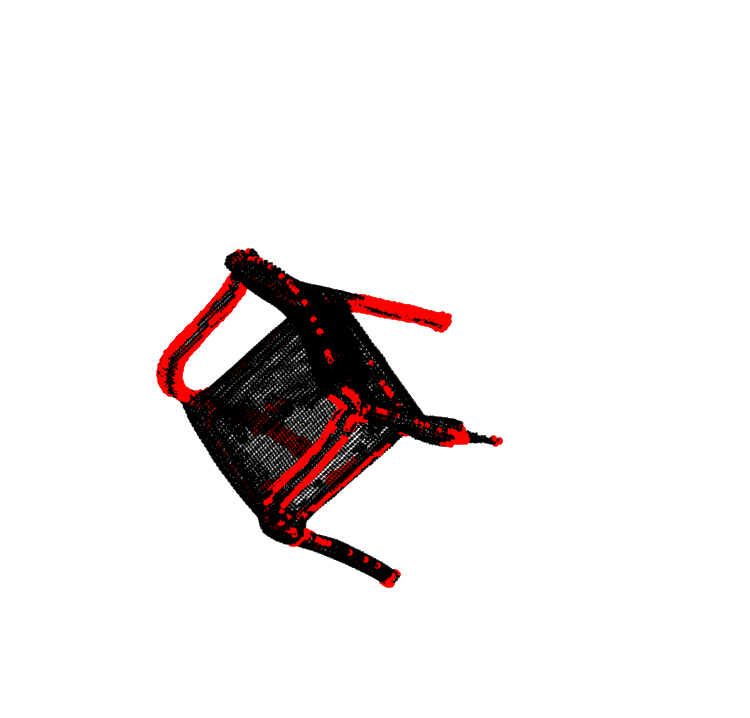}  
&
\includegraphics[width=0.45\columnwidth]{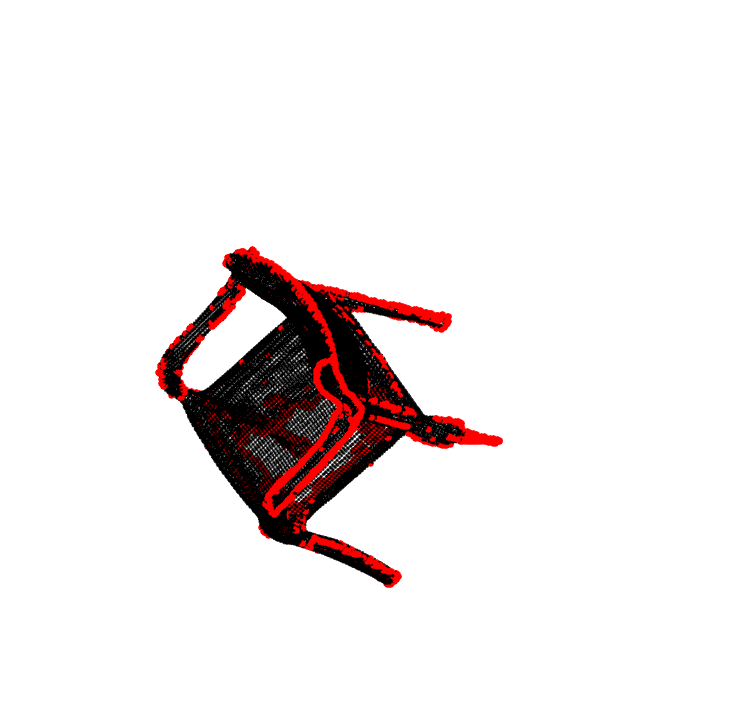}    
\\
{\small (c) Chair: Difference of normals. }  & {\small (d)  Chair: Local variation. }
\end{tabular}
  \end{center}
  \caption{\label{fig:contour_comparison} Haar-like high-pass graph filtering based local variation~\eqref{eq:LV} outperforms the DoN method.  }
  \vspace{-4mm}
\end{figure}

\begin{figure*}[htb]
  \begin{center}
    \begin{tabular}{lllll}
\includegraphics[width=0.3\columnwidth]{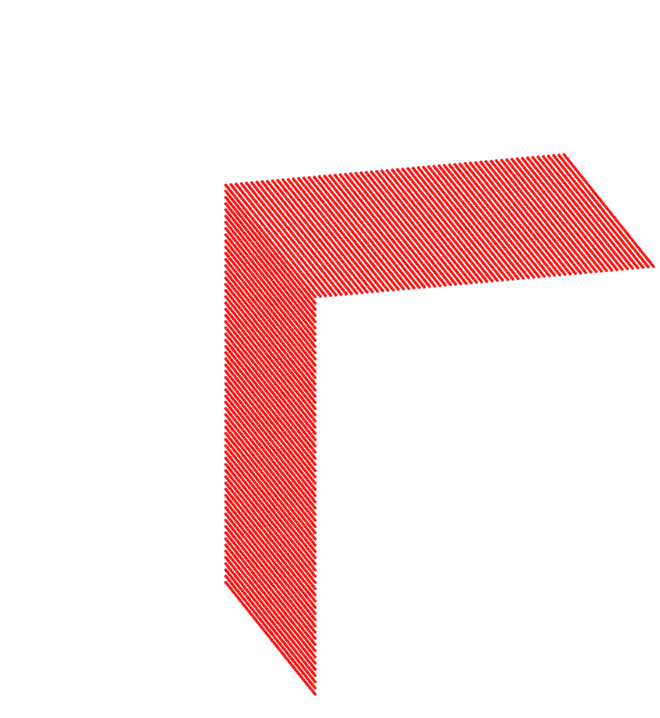}  
&
\includegraphics[width=0.28\columnwidth]{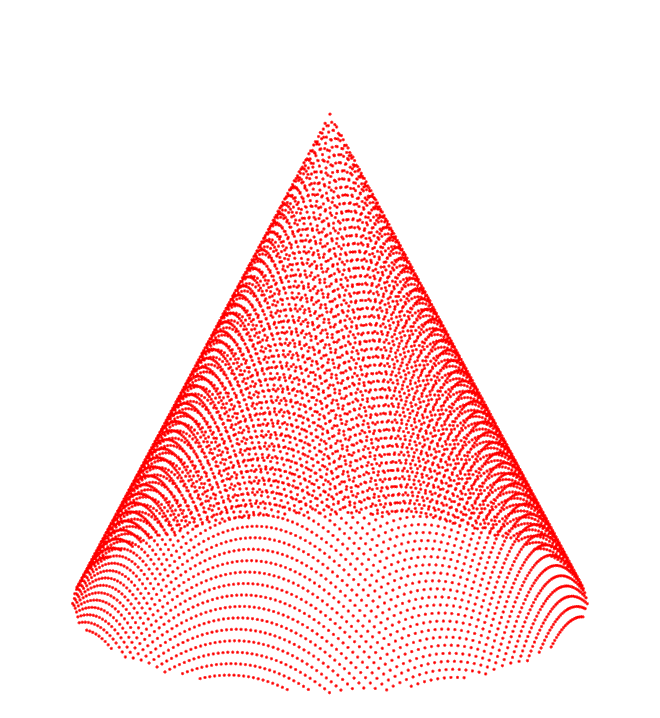}  
&
\includegraphics[width=0.33\columnwidth]{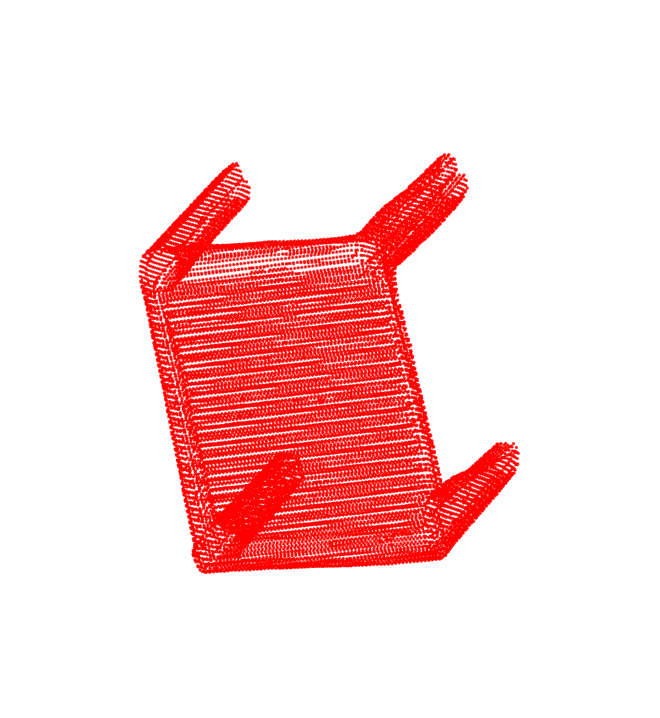}  
&
\includegraphics[width=0.32\columnwidth]{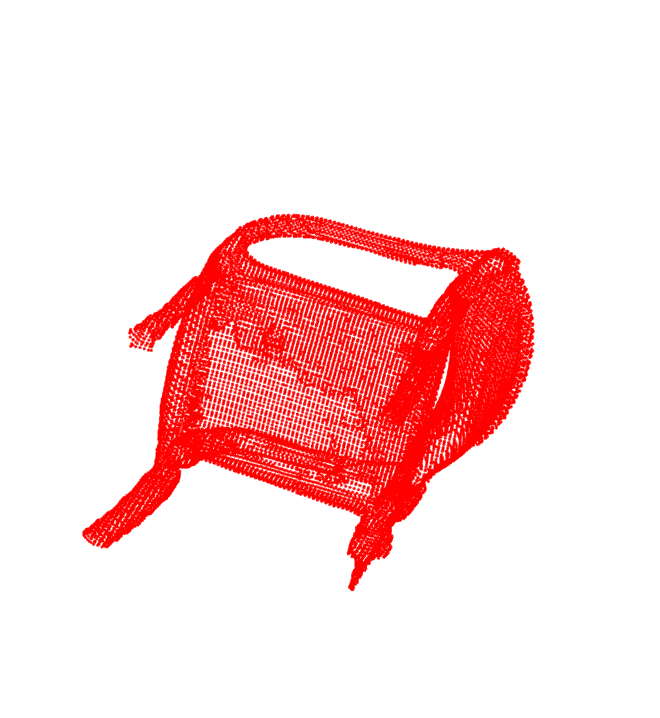}   
\\
\includegraphics[width=0.3\columnwidth]{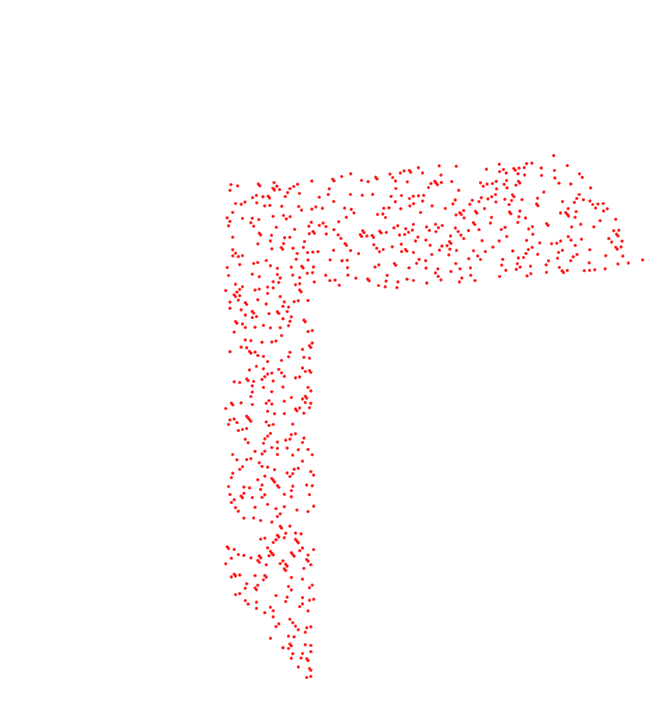}  
&
\includegraphics[width=0.28\columnwidth]{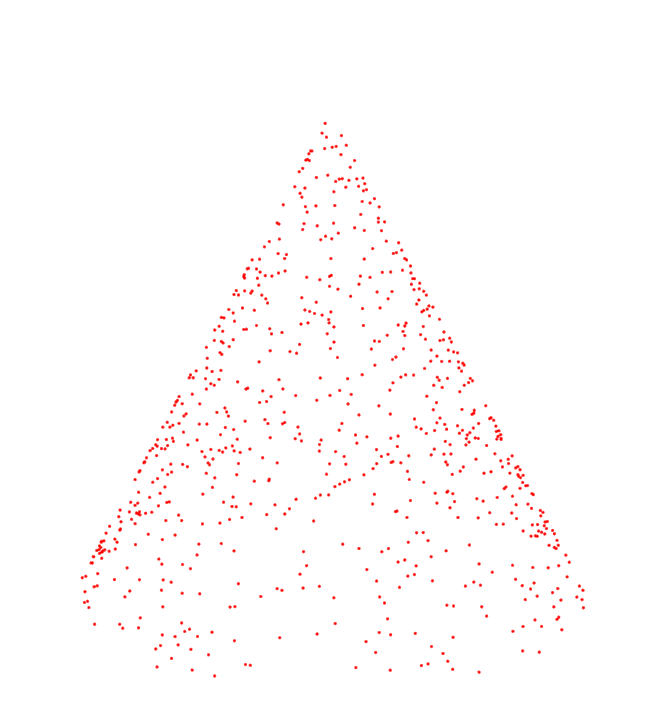}  
&
\includegraphics[width=0.33\columnwidth]{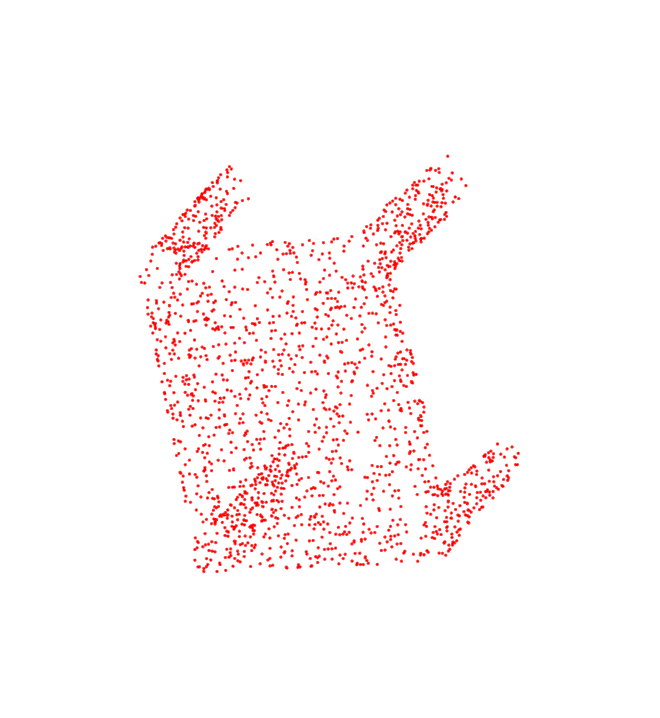}  
&
\includegraphics[width=0.33\columnwidth]{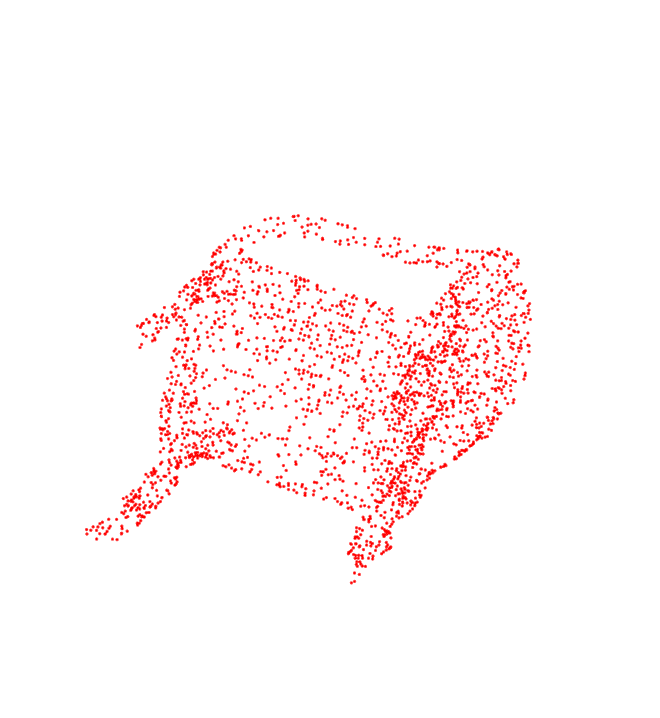}  
\\
\includegraphics[width=0.3\columnwidth]{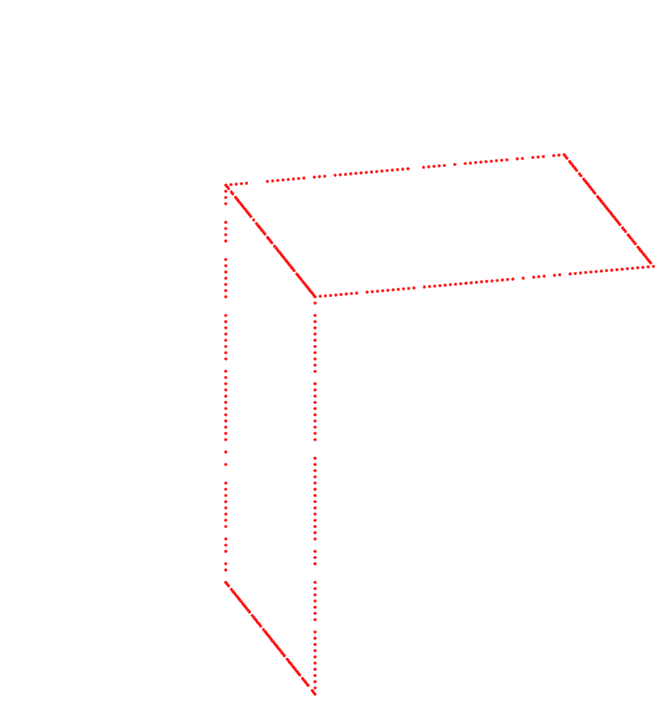}  
&
\includegraphics[width=0.28\columnwidth]{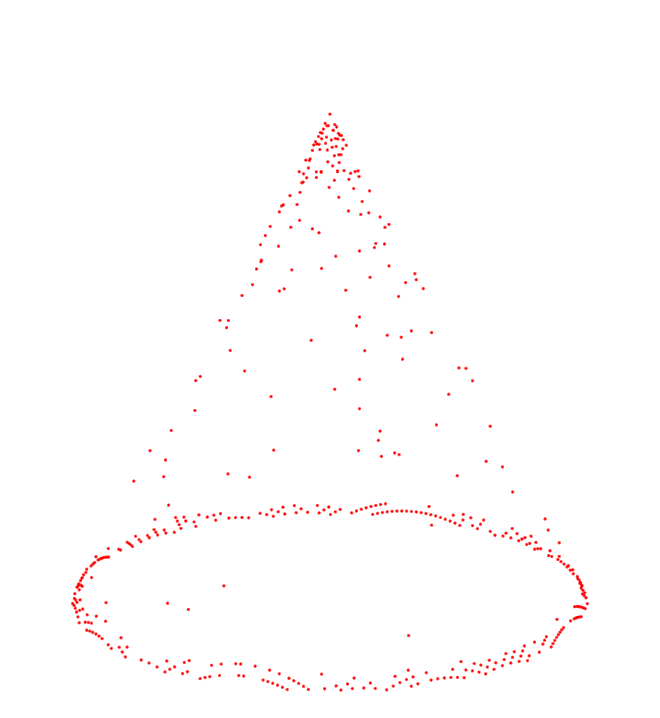}  
&
\includegraphics[width=0.33\columnwidth]{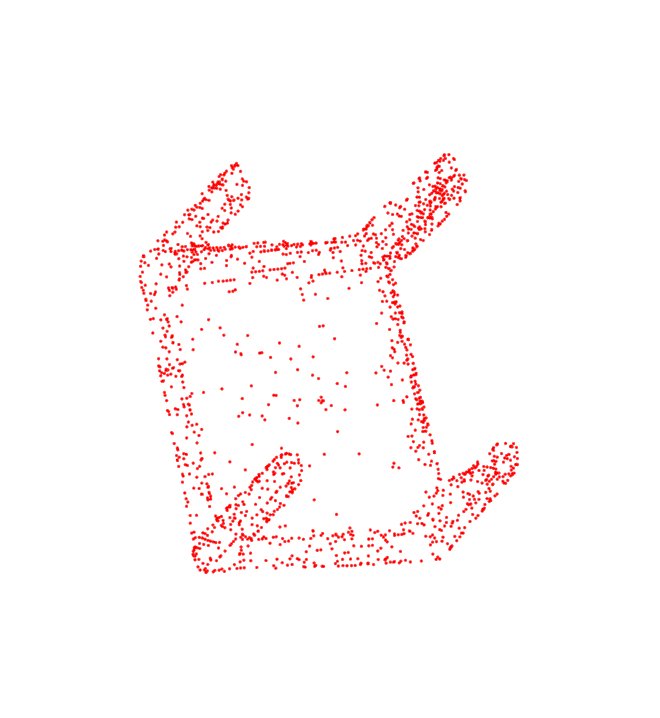}  
&
\includegraphics[width=0.33\columnwidth]{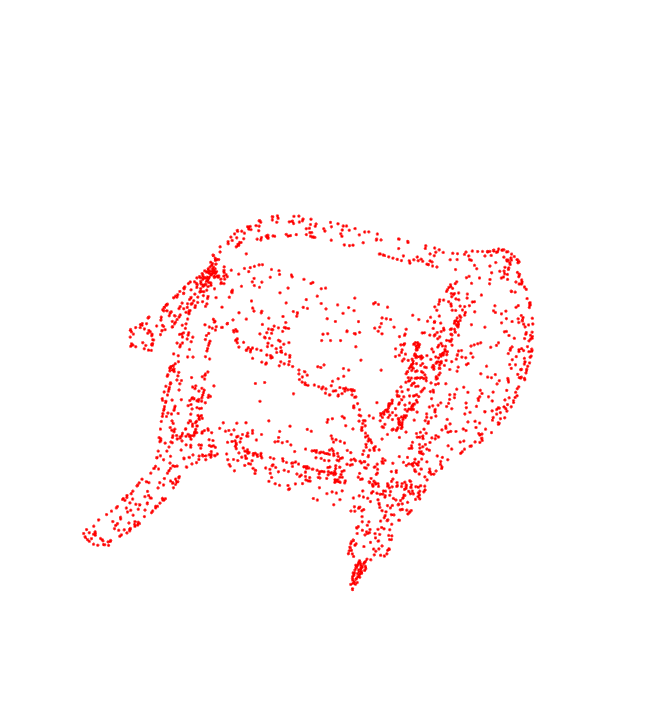}  
\end{tabular}
  \end{center}
  \caption{\label{fig:sample_scores}  Haar-like high-pass graph filtering based local variation~\eqref{eq:LV} outperforms pairwise difference based local variation~\eqref{eq:pair_LV}.  We use local variation to capture the contour. The first row shows the original point clouds; the second and third rows show the resampled versions with respect to two local variations: pairwise difference based local variation~\eqref{eq:pair_LV} and Haar-like high-pass graph filtering based local variation~\eqref{eq:LV}. Two resampled versions have the same number of points, which is $10\%$ of points in the original point cloud.}
\end{figure*}

\mypar{Experimental Validations}
Figure~\ref{fig:contour_comparison} compares the Haar-like high-pass graph filtering based local variation~\eqref{eq:LV} (second column) with that computed from the difference of normals (DoN) method (first column) ~\cite{LoannouTHG:12} which is used to analyze point clouds for segmentation and contour detection. As a contour detection technique, DoN computes the difference between surface normals calculated at two scales. In each plot, we highlight the points that have top $10\%$ largest DoN scores or local variations. In Figure~\ref{fig:contour_comparison} (a), we see that DoN cannot find the boundary in the plane because the surface normal does not change. The performance of DoN is also sensitive to predeisgned radius. For example, the difference of normals cannot capture precise contours in the hinge. On the other hand, local variation captures all the contours precisely in Figure~\ref{fig:contour_comparison} (b). We see similar results in Figures~\ref{fig:contour_comparison} (c), (d), (e) and (f). Further, difference of normals needs to compute the first principle component of the neighboring points for each 3D point, which is computationally inefficient. The local variation only involves a sparse matrix and vector multiplication, which is computationally efficient.

Figure~\ref{fig:sample_scores} shows the local variation based resampling distribution on some examples of the point cloud, including hinge, cone, table, chair and trash container. The first column shows the original point clouds; the second and third rows show the resampled versions with respect to two local variations: pairwise difference based local variation~\eqref{eq:pair_LV} and Haar-like high-pass graph filtering based local variation~\eqref{eq:LV}. Two resampled versions have the same number of points, which is $10\%$ of points in the original point cloud.

For two simulated objects,  the hinge and the cone (first two rows), the pairwise difference based local variation~\eqref{eq:pair_LV} fails to detect contour  and the Haar-like high-pass graph filtering based local variation~\eqref{eq:LV} detects all the contours. For the real objects, the Haar-like high-pass graph filtering based resampling~\eqref{eq:LV} also outperform the pairwise difference based local variation~\eqref{eq:pair_LV}. In summary, the Haar-like high-pass graph filtering based local variation~\eqref{eq:LV}  shows the contours of objects by using only $10\%$ of points.

\begin{figure}[htb]
  \begin{center}
    \begin{tabular}{cc}
\includegraphics[width=0.48\columnwidth]{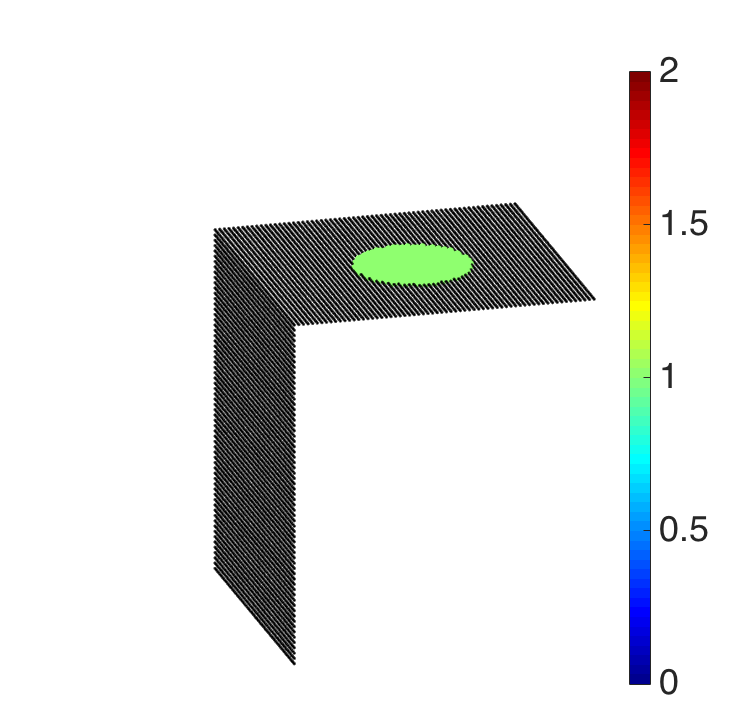}  
&
\includegraphics[width=0.42\columnwidth]{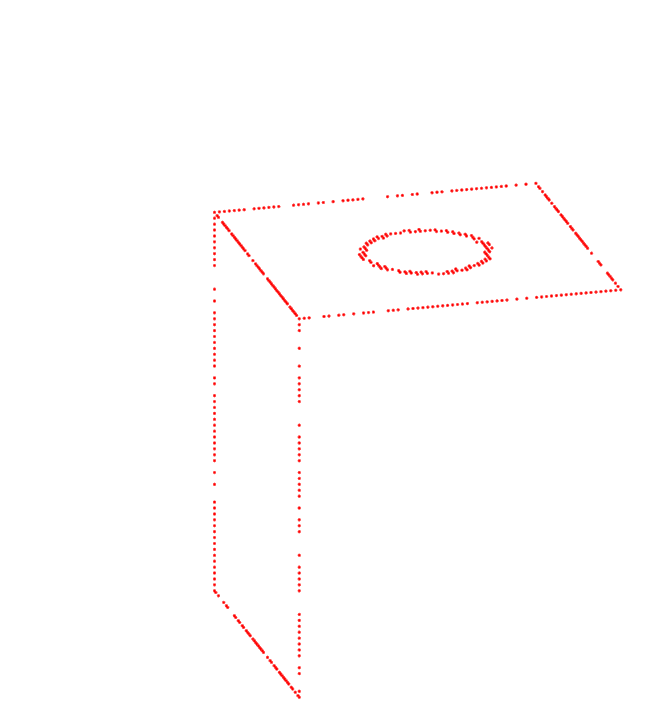} 
\\
{\small (a) Hinge with textures.}  & {\small (b) Resampled version.}
\\
\end{tabular}
  \end{center}
  \caption{\label{fig:twoplane_texture} High-pass graph filtering based resampling strategy detects both the geometric contour and the texture contour.}
\end{figure}

The high-pass graph filtering based resampling strategy can be easily extended to detect transient changes in other attributes. Figure~\ref{fig:twoplane_texture} (a) simulates a hinge with two different textures. The points in black have the same texture with value 0 and the points indicated by a green circle have a different texture with value 1. We put the texture as a new attribute and the point cloud matrix $\X \in \R^{N \times 4}$, where the first three columns are 3D coordinates and the fourth column is the texture. We resample $10\%$ of points based on the high-pass graph filtering based local variation~\eqref{eq:LV}. Figure~\ref{fig:twoplane_texture} (b) shows the resamped point cloud, which clearly detects both the geometric contour and the texture contour.

\begin{figure*}[thb]
  \begin{center}
    \begin{tabular}{ccccc}
\includegraphics[width=0.3\columnwidth]{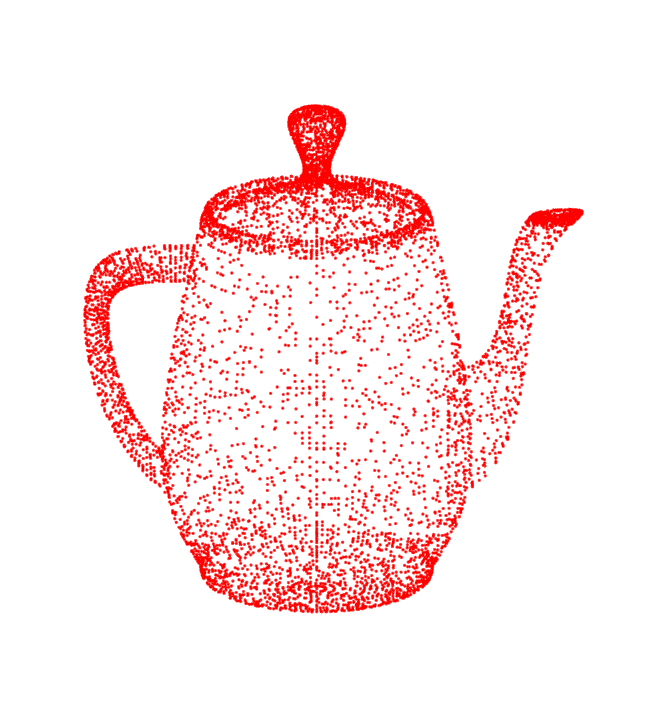}  
&
\includegraphics[width=0.28\columnwidth]{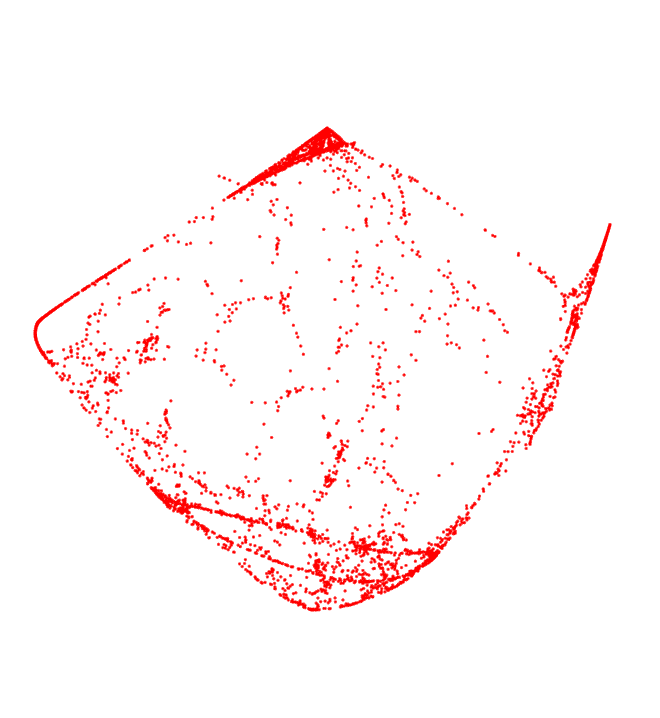} 
&
\includegraphics[width=0.3\columnwidth]{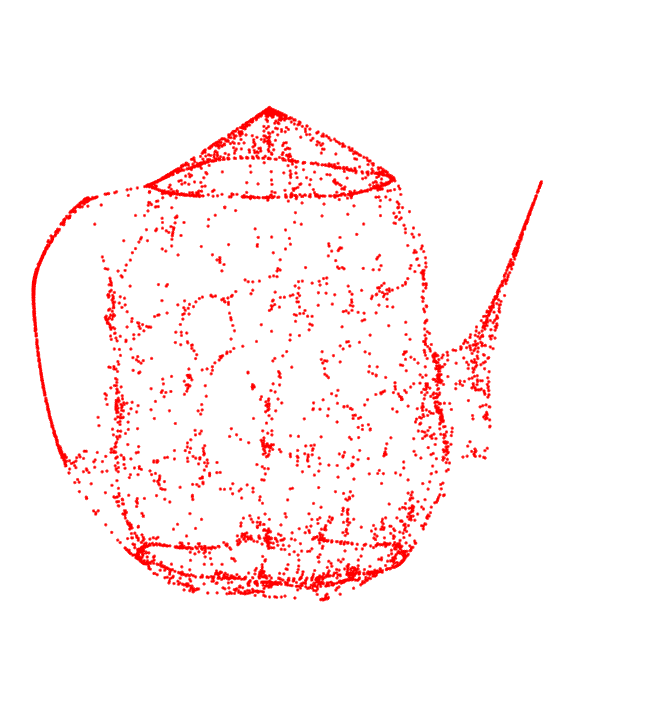} 
&
\includegraphics[width=0.3\columnwidth]{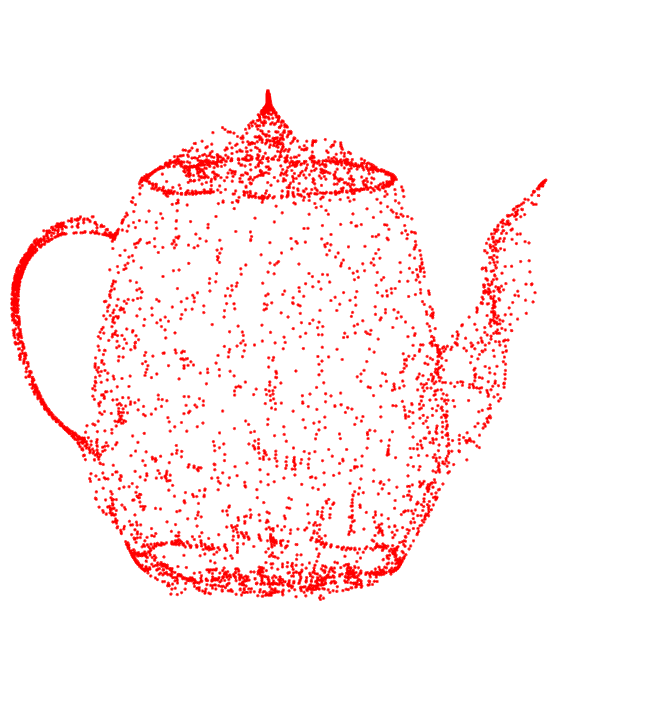} 
&
\includegraphics[width=0.4\columnwidth]{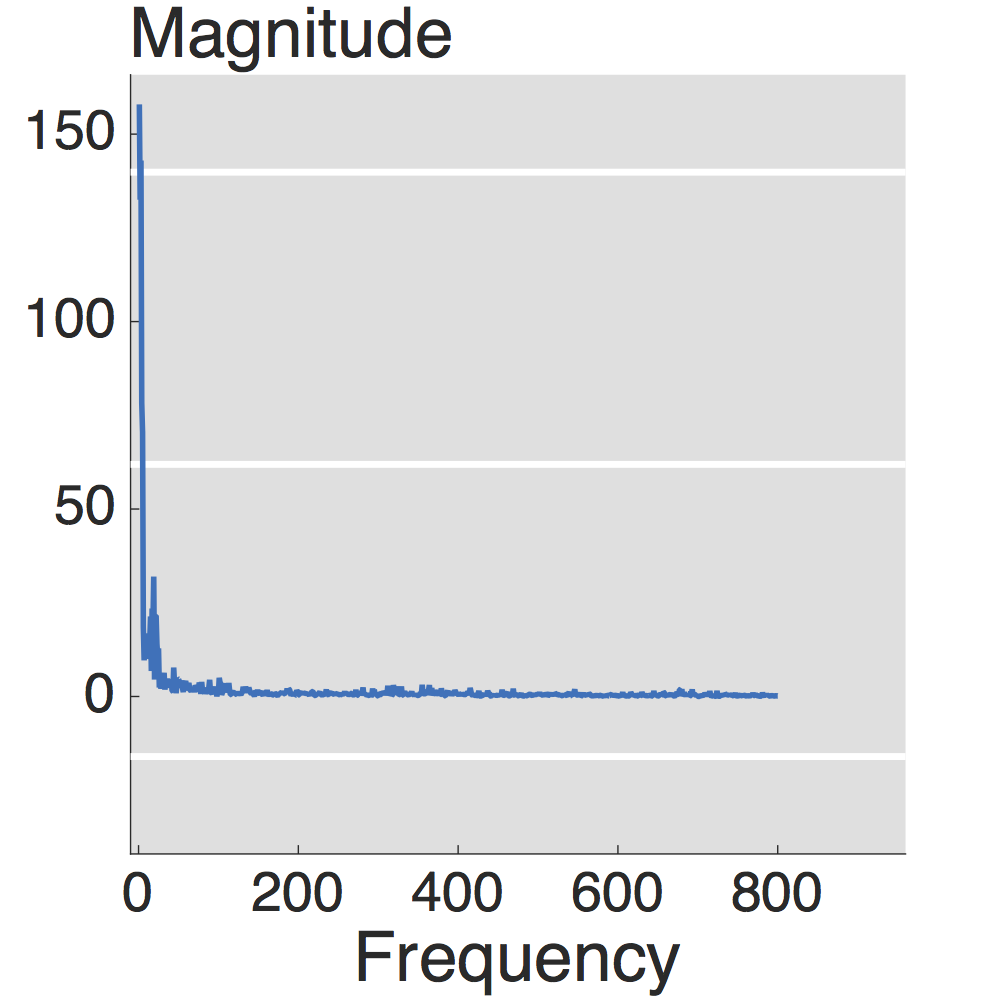} 
\\
{\small (a) Teapot. }  & {\small (b) Approximation with } & {\small (c) Approximation with } & {\small (d) Approximation with } & {\small (e) Graph spectral distribution.}
\\
&  {\small 10 graph frequencies. } & {\small   100 graph frequencies. } & {\small   500 graph frequencies. } & 
\\
\end{tabular}
  \end{center}
  \caption{\label{fig:teapot_example}  Low-pass approximation represents the main shape of the original point clouds. Plot (a) shows a point cloud with $8,000$ points representing a teapot. Plots (b), (c) and (d) show the approximations with $10$, $100$ and $500$ graph frequencies. We see that  the approximation with 10 graph frequencies shows a rough structure of a teapot; the approximation with 100 graph frequencies can be recognized as a teapot; the approximation with 500 graph frequencies show some details of the teapot. Plot (e) shows the graph spectral distribution, which clearly shows that most energy is concentrated in the low-pass band. }
\end{figure*}

\subsection{Low-pass Graph Filtering} 
In classical signal processing, a low-pass filter is used to capture rough shape of a smooth signal and reduce noise. Similarly, we use a low-pass graph filter to capture rough shape of a point cloud and reduce sampling noise during resampling.  Since we use the 3D coordinates of points to construct a graph~\eqref{eq:adj}, the 3D coordinates are naturally smooth on this graph, meaning that two adjacent points in the graph have similar coordinates in the 3D space. When a 3D point cloud is corrupted by noises and outliers, a low-pass graph filter, as a denoising operator, uses local neighboring information to approximate a true position for each point. Since the output after low-pass graph filtering is a denoised version of the original point cloud, it is more appropriate to resample from denoised points than original points.

\subsubsection{Ideal low-pass graph filter}
A straightforward choice is an ideal low-pass graph filter, which completely eliminates all graph frequencies above a given graph frequency while passing those below unchanged. An ideal low-pass graph filter with bandwidth $b$ is 
\begin{eqnarray*}
\vspace{-5mm}
h_{\rm IL}(\Adj)  & = &
\Vm  \bigg[ 
\begin{array}{ll}
\Id_{b \times b}  & \bold{0}_{b \times (N-b)} \\
\bold{0}_{(N-b) \times b} & \bold{0}_{(N-b) \times (N-b)}
\end{array} \bigg]
 \Vm^{-1}
 \\
& = &
 \Vm_{(b)} \Vm_{(b)}^T \in \R^{N \times N},
\end{eqnarray*}
where $\Vm_{(b)}$ is the first $b$ columns of $\Vm$,
and the graph frequency response is
\vspace{-4mm}
\begin{equation}
h_{\rm IL}(\lambda_i) \ = \ 
  \left\{ 
    \begin{array}{rl}
      1, & i \leq b;\\
      0, & \mbox{otherwise}.
  \end{array} \right. 
\end{equation}
The ideal low-pass graph filter $h_{\rm IL}$  projects an input graph signal onto a bandlimited subspace~\cite{ChenVSK:15} and $h_{\rm IL}(\Adj)  \s$ is a bandlimited approximation of the original graph signal $\s$. We show an example in Figure~\ref{fig:teapot_example}. Figure~\ref{fig:teapot_example} (b), (c) and (d) shows that the bandlimited approximation of the 3D coordinates of a teapot gets better when the bandwidth $b$ increases. We see that the bandwidth influences the shape of the teapot rapidly: with ten graph frequencies, we only obtain a rough structure of the teapot. Figure~\ref{fig:teapot_example} (e) shows that the main energy is concentrated in the low-pass graph frequency band.

The feature-extraction operator $f(\X) = \Vm_{(b)} \Vm_{(b)}^T \X$ is shift and rotation-varying. Based on Theorem~\ref{thm:opt_sample_variant}, the corresponding optimal resampling strategy is
\begin{eqnarray}
\label{eq:ideal_low_pass_opt_sample}
 \pi^*_i  & \propto  &  \sqrt{  c^2 \left\|  \left( \Vm_{(b)} \right)_i \right\|_2^2  +   \left\| \left( \Vm_{(b)} \Vm_{(b)}^T \Xo \right)_i \right\|_2^2 }
 \\
 & = &
 \nonumber
  \sqrt{  c^2 \left\|  \vv_i \right\|_2^2  +   \left\|   \Xo^T \Vm_{(b)}  \vv_i \right\|_2^2 },
\end{eqnarray}
where $\vv_i \in \R^b$ is the $i$th row of $\Vm_{(b)}$.

A direct way to obtain $\left\| \vv_i \right\|_2$ requires the truncated eigendecomposition~\eqref{eq:graph_FT}, whose computational cost is $O(Nb^2)$, where $b$ is the bandwidth. It is potentially possible to approximate the leverage scores through a fast algorithm~\cite{DrineasMMW:12, Woodruff:14}, where we use randomized techniques to avoid the eigendecomposition and the computational cost is $O(Nb \log(N))$. Another way to leverage computation is to partition a graph into several subgraphs and obtain leverage scores in each subgraph.

%Note that this resampling strategy is similar to sampling and recovery of approximately bandlimited graph signals~\cite{ChenVSK:15c}, whose idea is to sample the signal coefficients at a few nodes and approximately recover the signal coefficients at all the other nodes. Here we model the attributes of the point cloud as graph signals, sample the attributes of a few points and approximately recover the attributes of all the other points.

%\begin{figure}[htb]
%  \begin{center}
%    \begin{tabular}{cc}
%\includegraphics[width=0.4\columnwidth]{figures/bunny/noisy.png}   & \includegraphics[width=0.4\columnwidth]{figures/bunny/noisy_sample.png} 
%\\
%{\small (a) Noisy bunny. }  & {\small (b) Uniform resampling.  }
%\\
%\includegraphics[width=0.4\columnwidth]{figures/bunny/denoise.png}   & \includegraphics[width=0.4\columnwidth]{figures/bunny/denoise_sample.png} 
%\\
%{\small (c) Denoised bunny. }  & {\small (d) Resampling based on }
%\\
% & {\small Haar-like low-pass graph filtering. }
%\\
%\end{tabular}
%  \end{center}
%  \caption{\label{fig:haar_sample_scores}  Resampling based on Haar-like low-pass graph filtering is robust to noise. Plot (a) shows a noisy point cloud of bunny contaminated by a Gaussian noise. Plot (b) shows samples that are uniformly sampled from the noisy bunny. Plot (c) shows a denoised version of bunny by using the Haar-like graph filtering.   Plot (d) shows samples selected by ~\eqref{eq:haar_opt_sample}. }
% \end{figure}

\subsubsection{Haar-like low-pass graph filter}
\label{sec:haar_low_filter}
Another simple choice is Haar-like low-pass graph filter; that is,
\begin{eqnarray}
\label{eq:HL}
 && h_{\rm HL}(\Adj) 
\ = \   \Id + \frac{1}{|\lambda_{\rm max}|} \Adj  
\\
\nonumber
& = &  \Vm 
\begin{bmatrix}
1 +  \frac{\lambda_1}{|\lambda_{\rm max}|} &  0 & \cdots & 0  \\
0  &  1 + \frac{\lambda_2}{|\lambda_{\rm max}|}  & \cdots & 0 \\
\vdots &  \vdots & \ddots & \vdots  \\
0  &  0 & \cdots & 1 + \frac{\lambda_N}{|\lambda_{\rm max}|} 
\end{bmatrix}
 \Vm^{-1},
\end{eqnarray}
where $\lambda_{\rm max} = \max_{i} |\lambda_i|$ with $\lambda_i$ eigenvalues of $\Adj$. The normalization factor $\lambda_{\rm max}$ is presented to avoid the amplification of the magnitude. We denote $ \Adj_{\rm norm} =  \Adj /|\lambda_{\rm max}|$ for simplicity. The graph frequency response is $h_{\rm HL}(\lambda_i) = 1+ \lambda_i/|\lambda_{\rm max}|$. Since the eigenvalues are ordered in a descending order, we have $1+ \lambda_i \geq 1+ \lambda_{i+1}$, meaning low frequency response amplifies and high frequency response attenuates.

In the graph vertex domain, the response of the $i$th point is
$
\left( h_{\rm HL}(\Adj)  \X \right)_i \ = \  \x_i +  \sum_{j \in \N_i} (\Adj_{\rm norm})_{i,j} \x_j,
$
where $\N_i$ is the neighbors of the $i$th point. We see that  $h_{\rm HL}(\Adj)$ averages the attributes of each point and its neighbors to provide a smooth output. 

The feature-extraction operator $f(\X) = h_{\rm HL} (\Adj) \X$ is shift and rotation-variant. Based on Theorem~\ref{thm:opt_sample_variant}, the corresponding optimal resampling strategy is
\begin{eqnarray}
\label{eq:haar_opt_sample}
 \pi^*_i  & \propto  &  \sqrt{  c^2 \left\|  \left( \Id + \Adj_{\rm norm} \right)_i \right\|_2^2  +   \left\| \left( \left(\Id + \Adj_{\rm norm} \right) \Xo \right)_i \right\|_2^2 },
\nonumber \\
\end{eqnarray}

To obtain this optimal resampling distribution,  we need to compute the largest magnitude eigenvalue $\lambda_{\rm max}$, which takes $O(N)$, and compute $\left\| \left( \Id + \Adj_{\rm norm} \right)_i \right\|_2^2$  and $\left\| \left( \left(\Id + \Adj_{\rm norm} \right) \Xo \right)_i \right\|_2^2$ for each row, which takes $O( \left\| {\rm vec} (\Adj) \right\|_0)$ with $\left\| {\rm vec} (\Adj) \right\|_0$ the nonzero elements in the graph shift operator. We can avoid computing the largest magnitude by using a normalized adjacency matrix or a transition matrix as a graph shift operator.  A normalized adjacency matrix is $\D^{-\frac{1}{2}} \W \D^{-\frac{1}{2}}$, where $\D$ is the diagonal degree matrix, and a transition matrix is obtained by normalizing the sum of each row of an adjacency matrix to be one; that is $\D^{-1} \W$. In both cases, the largest eigenvalue of a transition matrix is one, we thus have $\Adj = \Adj_{\rm norm}$.

\begin{figure}[htb]
  \begin{center}
  \begin{tabular}{cc}
\includegraphics[width=0.28\columnwidth]{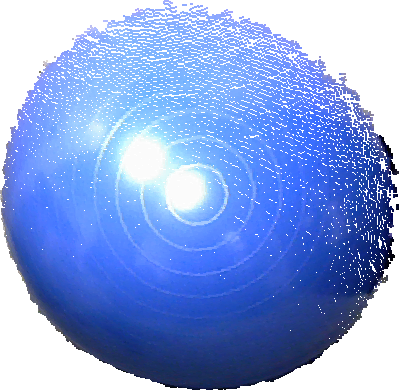}   
  & 
\includegraphics[width=0.28\columnwidth]{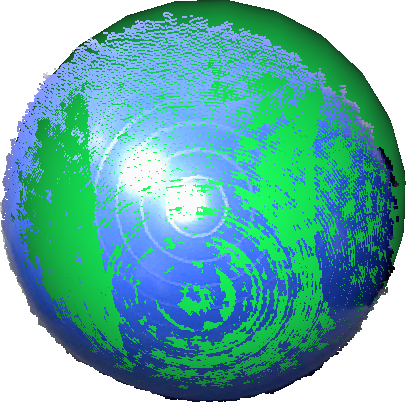} 
\\
{\small (a) Fitness ball. } & {\small (b) Sphere Fitting. }
  \end{tabular} 
  \end{center}
  \caption{\label{fig:fitness_ball_fit}  Shape modeling for a fitness ball.
}
\end{figure}

\begin{table*}[t]
  \footnotesize
  \begin{center}
    \begin{tabular}{@{}llllll@{}}
      \toprule
       &  Original ball &  Noisy ball  &  Uniform resampling &  Denoised ball & Low-pass graph filtering based resampling \\
      &   (Figure~\ref{fig:fitness_ball_fit} (a) ) & (Figure~\ref{fig:fitness_ball} (a) ) & (Figure~\ref{fig:fitness_ball} (b) ) & (Figure~\ref{fig:fitness_ball} (c) )  &  (Figure~\ref{fig:fitness_ball} (d) )    \\
      \midrule \addlinespace[1mm]
      Radius &  $0.3182$ & $0.3478 ~(9.3023\%)$  & $0.3520 (10.6223\%)$  & $0.3143 ~(1.2256\%)$ &   ${\bf 0.3199~(0.5343\%)}$ \\  
      Center-$x$ &  $0.0833$  & $0.0903 ~(8.4034\%)$  & $0.0975  ~(17.0468\%)$  & $0.0799 ~(4.0816\%)$ &  ${\bf 0.0849 ~(1.9208\%)}$ \\ 
      Center-$y$ &   $0.1903$  & $0.2136 ~(12.2438\%)$  & $0.1794 ~(5.7278\%)$  & ${\bf  0.1866 ~(1.9443\%)}$ &  $0.1783 ~(6.3058\%)$ \\ 
     Center-$z$ &   $1.1725$  & $1.3803 ~(17.7228\%)$  & $1.1530 ~(1.6631\%)$  & ${\bf 1.1618 ~(0.9126\%)}$ &  $1.1613 ~(0.9552\%)$ \\  
      \bottomrule
    \end{tabular}
  \end{center}
  \caption{\label{table:fitness_ball} Proposed resampling strategy with low-pass graph filtering provides a robust shape modeling for a fitness ball. The first column is the ground truth. The relative error is shown in the parentheses. Best results are marked in bold.  }
\end{table*}

\mypar{Experimental Validations} We aim to use a low-pass graph filter to handle a noisy point cloud. Figure~\ref{fig:fitness_ball_fit} (a) shows a  point cloud of a fitness ball, which contains $62,235$ points collected from a Kinect device. In this noiseless case, the surface of the fitness can be modeled by a sphere. Figure~\ref{fig:fitness_ball_fit} (b) fits a green sphere to the fitness ball~\footnote{Figure~\ref{fig:fitness_ball_fit} (b)  is generated from a public software CloudCompare.}. The radius and the central point of this sphere is $0.318238$ and $\begin{bmatrix}
0.0832627 &  0.190267 & 1.1725 \end{bmatrix}$. To leverage the computation, we resample a subset of points and fit another sphere to the resample points. We want these two spheres generated by the original point cloud and the resampled point cloud to be similar. 

%\Note{TODO: Let's try one more experiment: let's fit a sofa to a T-spline, I have the unorganized point cloud to T-spline fitting function now. Let's discuss later.}

\begin{figure}[htb]
  \begin{center}
    \begin{tabular}{cc}
    \vspace{-5mm}
\includegraphics[width=0.45\columnwidth]{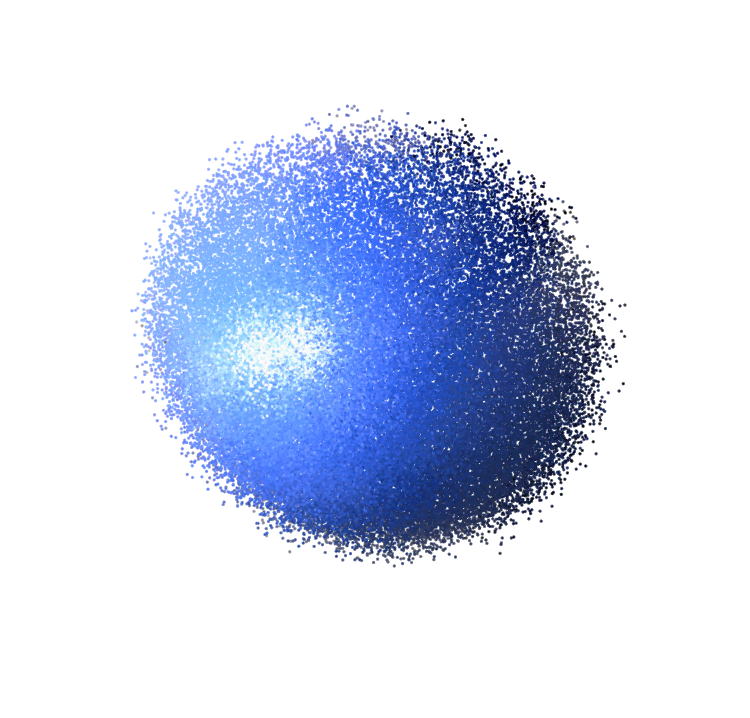}  
&
 \includegraphics[width=0.45\columnwidth]{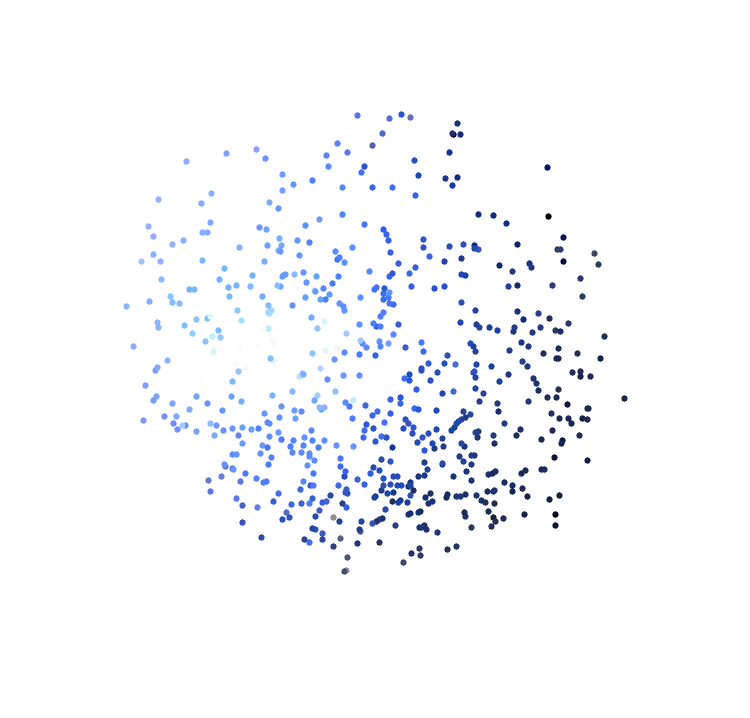}  
\\
 {\small (a) Noisy ball. } &  {\small (b) Uniform resampling.} 
\\
\includegraphics[width=0.45\columnwidth]{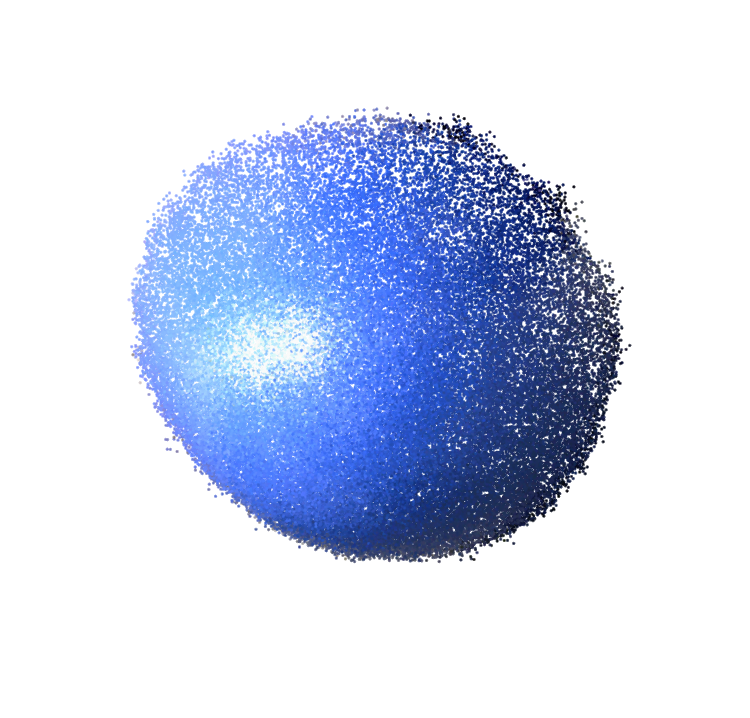} 
&
\includegraphics[width=0.45\columnwidth]{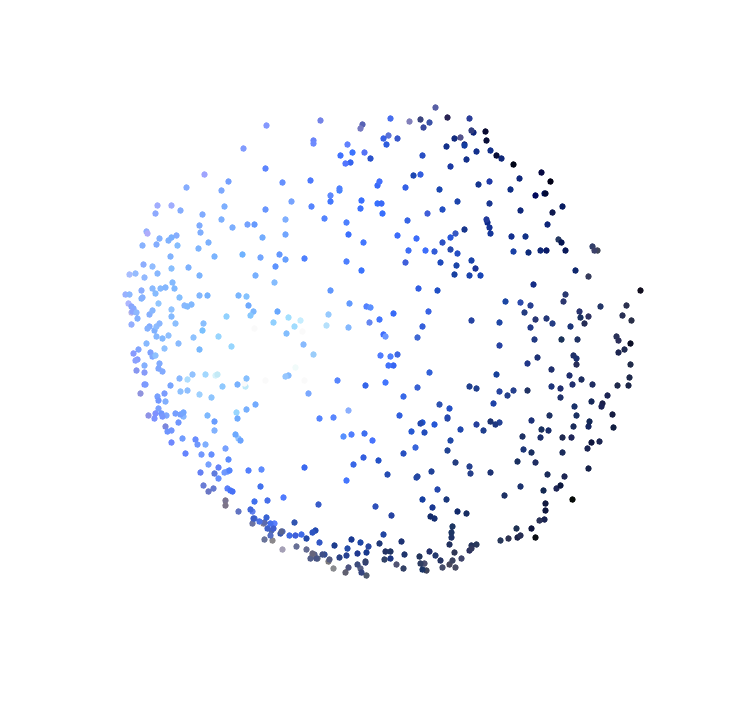}  
\\
 {\small (c) Denoised ball. } &  {\small (d)  Low-pass graph filtering }
 \\
 &  {\small based resampling. }
\\
\end{tabular}
  \end{center}
  \caption{\label{fig:fitness_ball} Denoising and resampling of a noisy fitness ball. Plot (c) denoises Plot (a). Plot (d) resamples from Plot (c)  according to the resampling strategy~\eqref{eq:ideal_low_pass_opt_sample}  }
\end{figure}

In many real cases, the original points are collected with noise. To simulate the noisy case, we add the Gaussian noise with mean zeros and variance $0.02$ to each points. Figures~\ref{fig:fitness_ball} (a) and (b) show a noisy point cloud and its resampled version based on uniform resampling, respectively. Figures~\ref{fig:fitness_ball} (c) and (d) show a denoised point cloud and its resampled version, respectively. The denoised point cloud is obtained by the low-pass graph filtering~\eqref{eq:HL} and the resampling strategy is  based on~\eqref{eq:ideal_low_pass_opt_sample}. We fit a sphere to each of the four point clouds and the statistics are shown in Table~\ref{table:fitness_ball}. The relative errors are shown in the parenthesis, which is defined as
$
{\rm Error} \ = \  \arrowvert (x - \hat{x})/x  \arrowvert,
$
where $x$ is the ground truth, and $\hat{x}$ is the estimation. The denoised ball and its resampled version outperform the noisy ball and the uniform-sampled version because the estimated radius and the central point is closer to the original radius and the central point. This validates that the proposed resampling strategy with low-pass graph filtering provides a robust shape modeling for noisy point clouds.

\begin{figure}[htb]
  \begin{center}
\includegraphics[width=0.8\columnwidth]{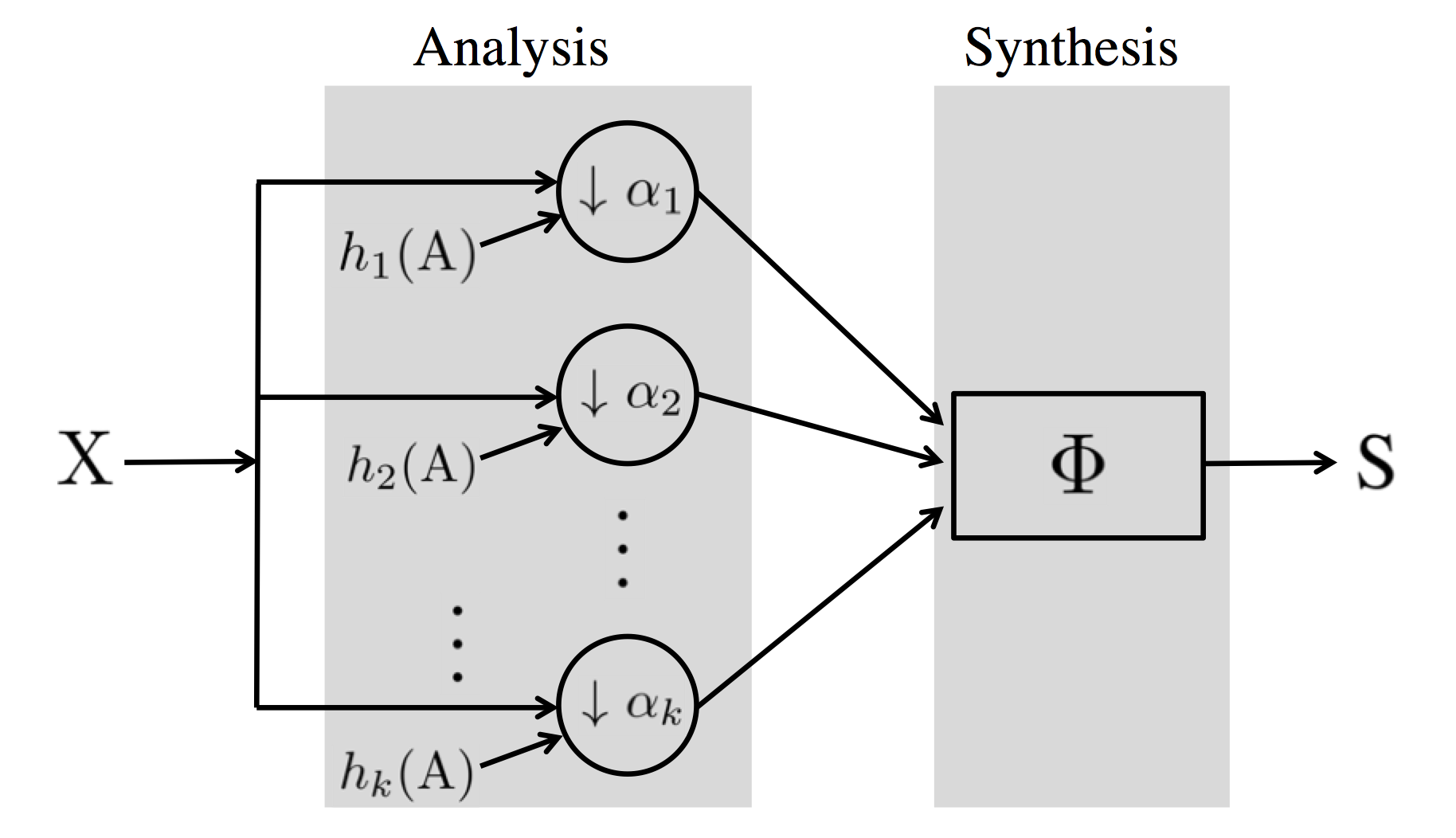} 
  \end{center}
  \caption{\label{fig:GFB}  Graph filter bank analysis for 3D point clouds. In the analysis part, we separate a 3D point cloud into multiple subbands. In each subband, we resample a subset of 3D points based on a specific graph filter $h(\Adj)$. The number of samples in each subband is determined by a sampling ratio $\alpha$.  In the synthesis part, we use all the resampled points to reconstruct a surface via a reconstruction operator $\Phi$.  }
\end{figure}

\subsection{Graph Filter Banks}
\label{sec:so_beautiful}
In classical signal processing,  a filter bank is an array of band-pass filters that analyze an input signal in multiple subbands and synthesize the original signal from all the subbands~\cite{VetterliKG:12,KovacevicGV:12}. We use a similar idea to analyze a 3D point cloud: separate an input 3D point cloud into multiple components via different resampling operators, allowing us to enhance different components of a 3D point cloud. For example, we resample both contour points and noncontour points to reconstruct the original surfaces, but we need more contour points to emphasize contours.

Figure~\ref{fig:GFB} shows a surface reconstruction system for a 3D point cloud based on graph filter banks. In the analysis part, we separate a 3D point cloud $\X$ into $k$ subbands. In each subband, the information preserved is determined by a specific graph filter and we resample a subset of 3D points according to~\eqref{eq:opt_sample_inv}  and~\eqref{eq:opt_sample_variant}. The number of samples in each subband is determined by a sampling ratio $\alpha$. We have flexibility to use either the original 3D points or the 3D points after graph filtering. In the synthesis part, we use the resampled points to reconstruct the surface. A literature review on surface reconstruction algorithms is shown in~\cite{BergerTSAGLSS:16}.  Since each surface reconstruction algorithm has its own specific set of assumptions, different surface reconstruction algorithms perform differently on the same set of 3D points.

We measure the overall performance of a surface reconstruction system by reconstruction error, which is the difference between the surface reconstructed from resampled points and the original surface. This leads to a rate-distortion like tradeoff: when we resample more points, we encode more bits and the reconstruction error is smaller; when we resample fewer points, we encode less bits and the reconstruction error is larger.  The overall goal is: given an arbitrary tolerance of reconstruction error, we use as few samples as possible to reconstruct a surface by carefully choosing a graph filter and sampling ratio in each subband.
Such a surface reconstruction system will benefit a 3D point cloud storage and compression because we only need to store a few resampled points. Since a surface reconstruction system is application-dependent, the design details are beyond the scope of this paper.

\section{Applications}
\label{sec:app}

\begin{figure*}[tb]
  \begin{center}
    \begin{tabular}{ccc}
\includegraphics[width=0.5\columnwidth]{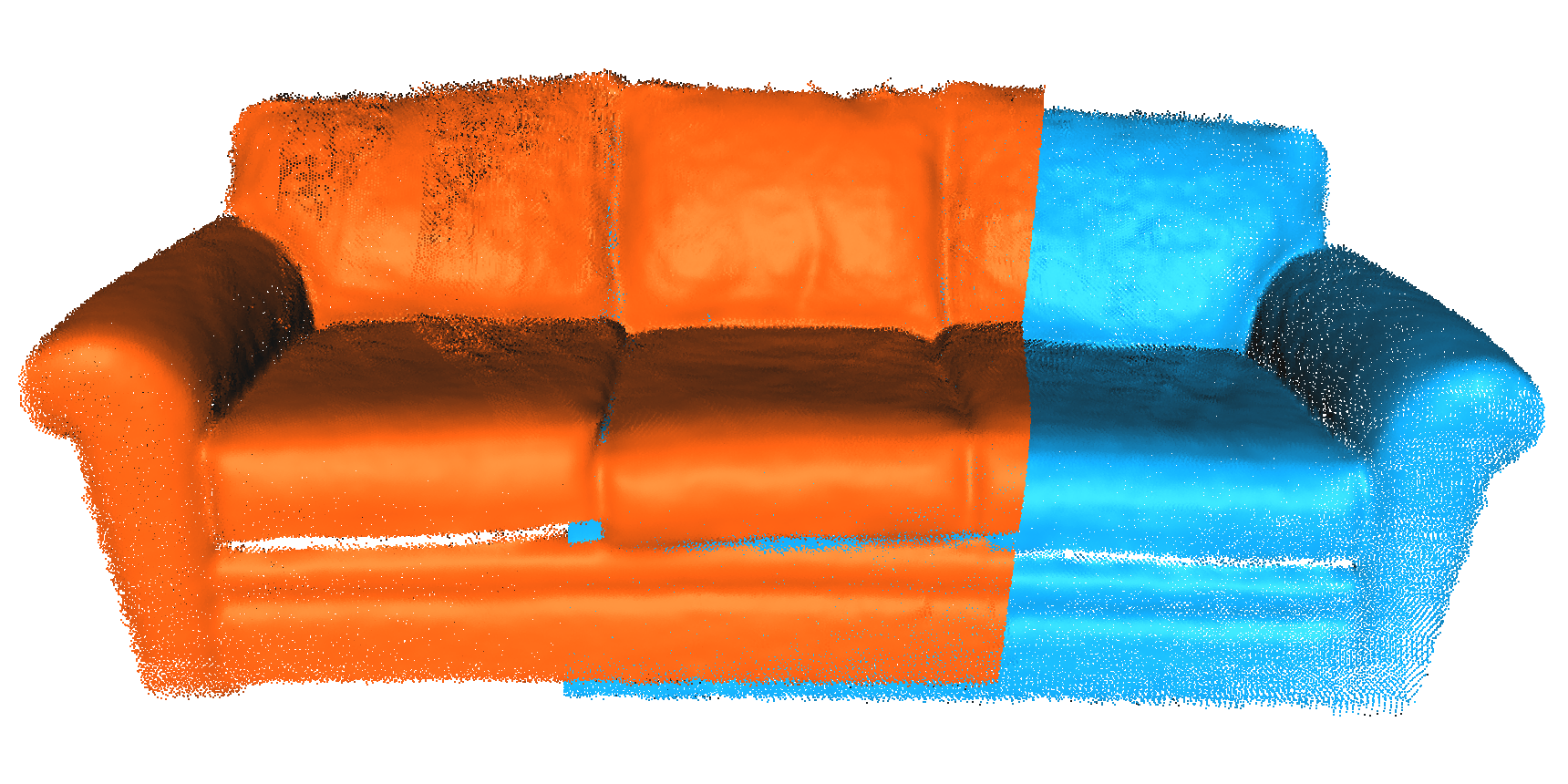}  
&
\includegraphics[width=0.5\columnwidth]{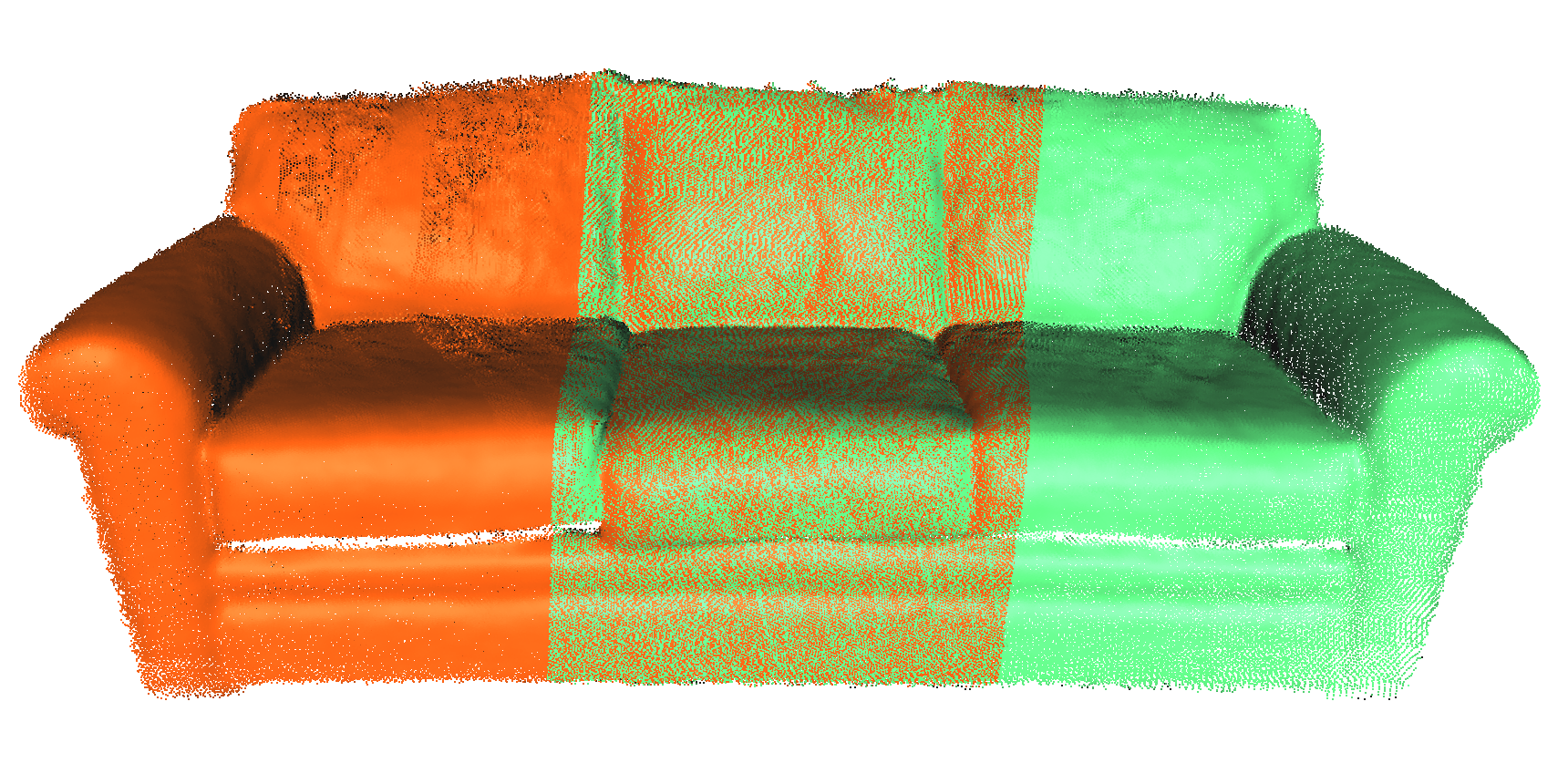} & 
\includegraphics[width=0.5\columnwidth]{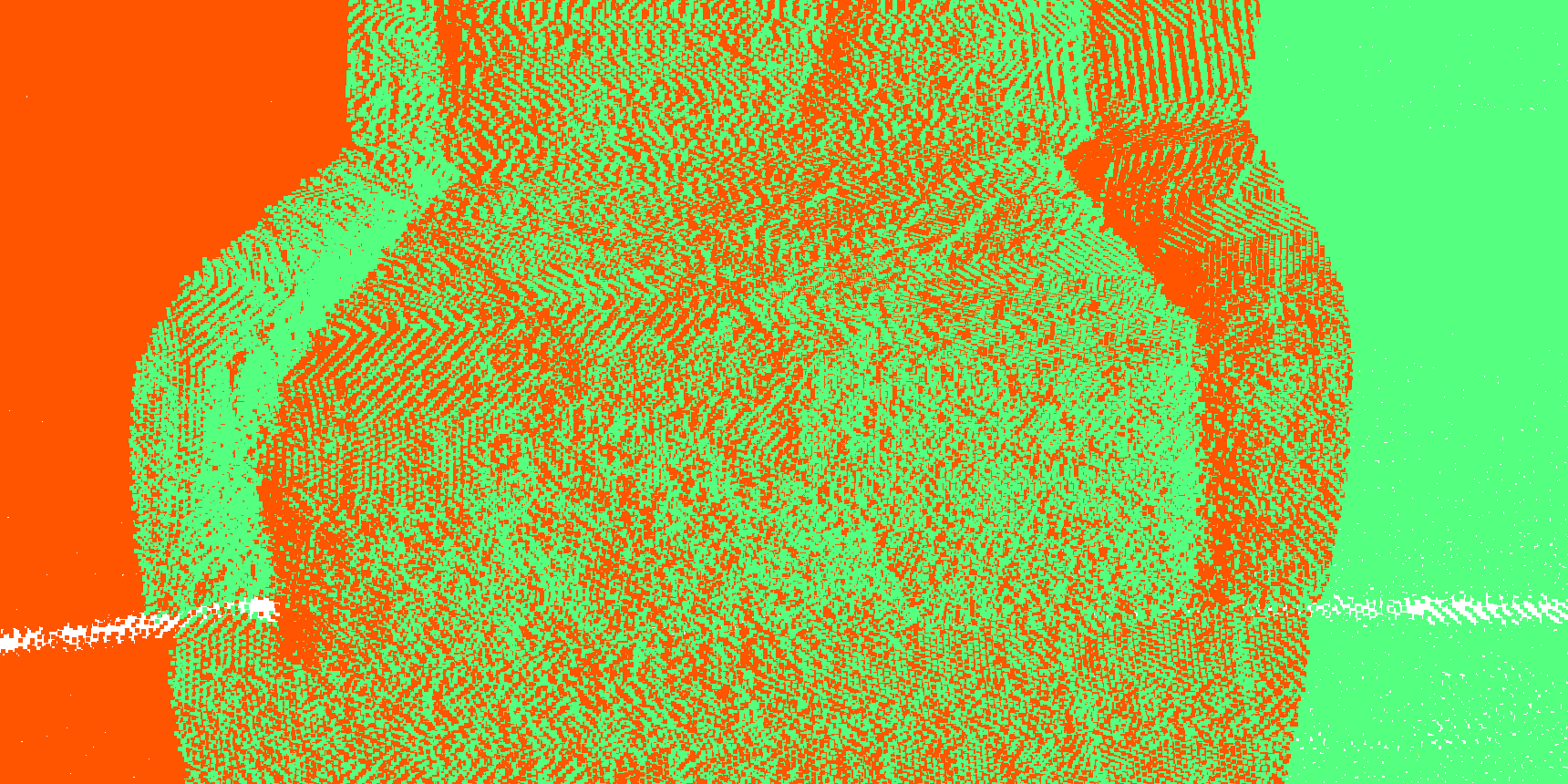}
\\
{\small (a) Original point cloud. }  &  {\small (b) Registered point cloud. }  &  {\small (c) Details. }
\\
\includegraphics[width=0.5\columnwidth]{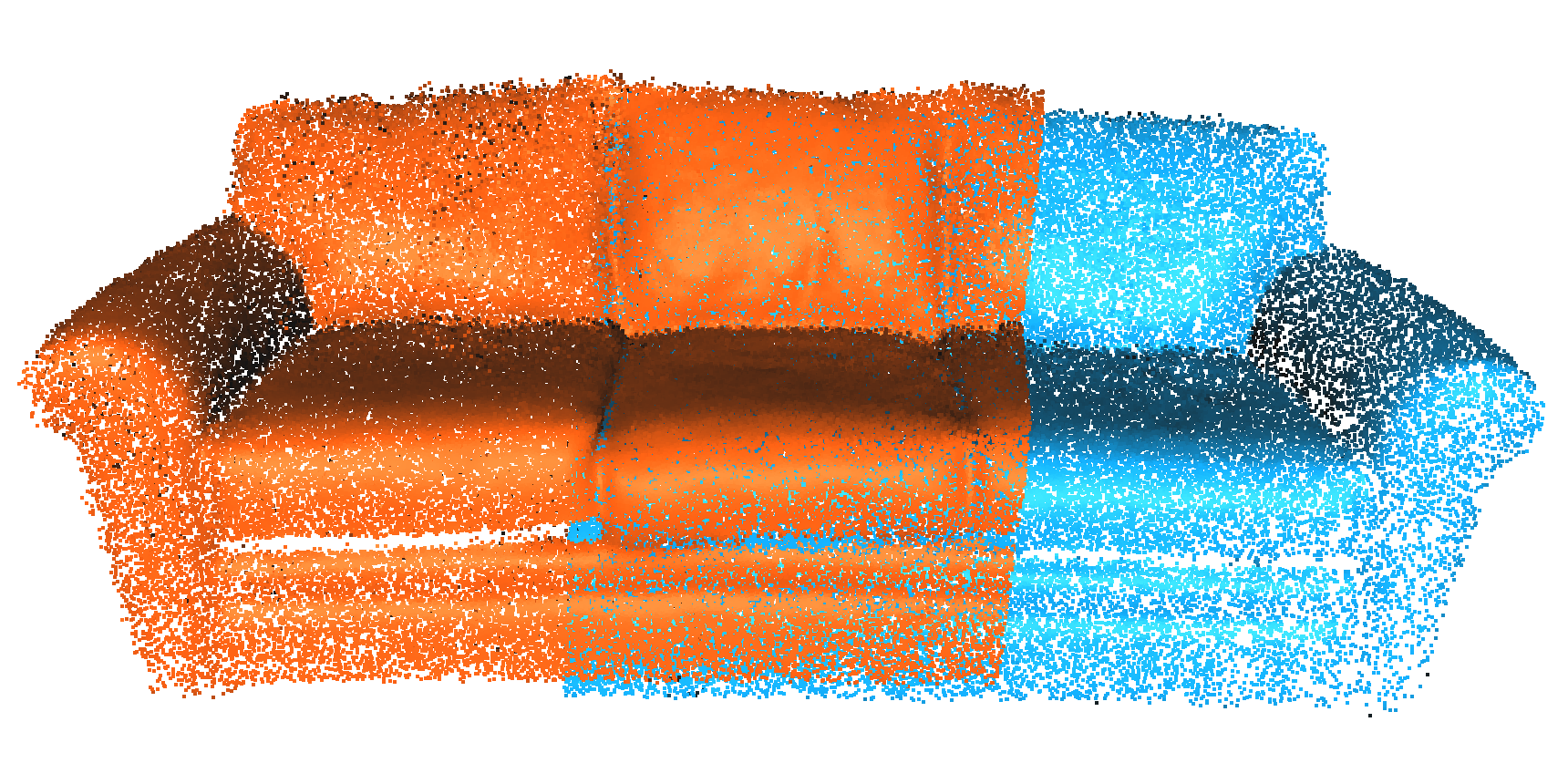}
&
\includegraphics[width=0.5\columnwidth]{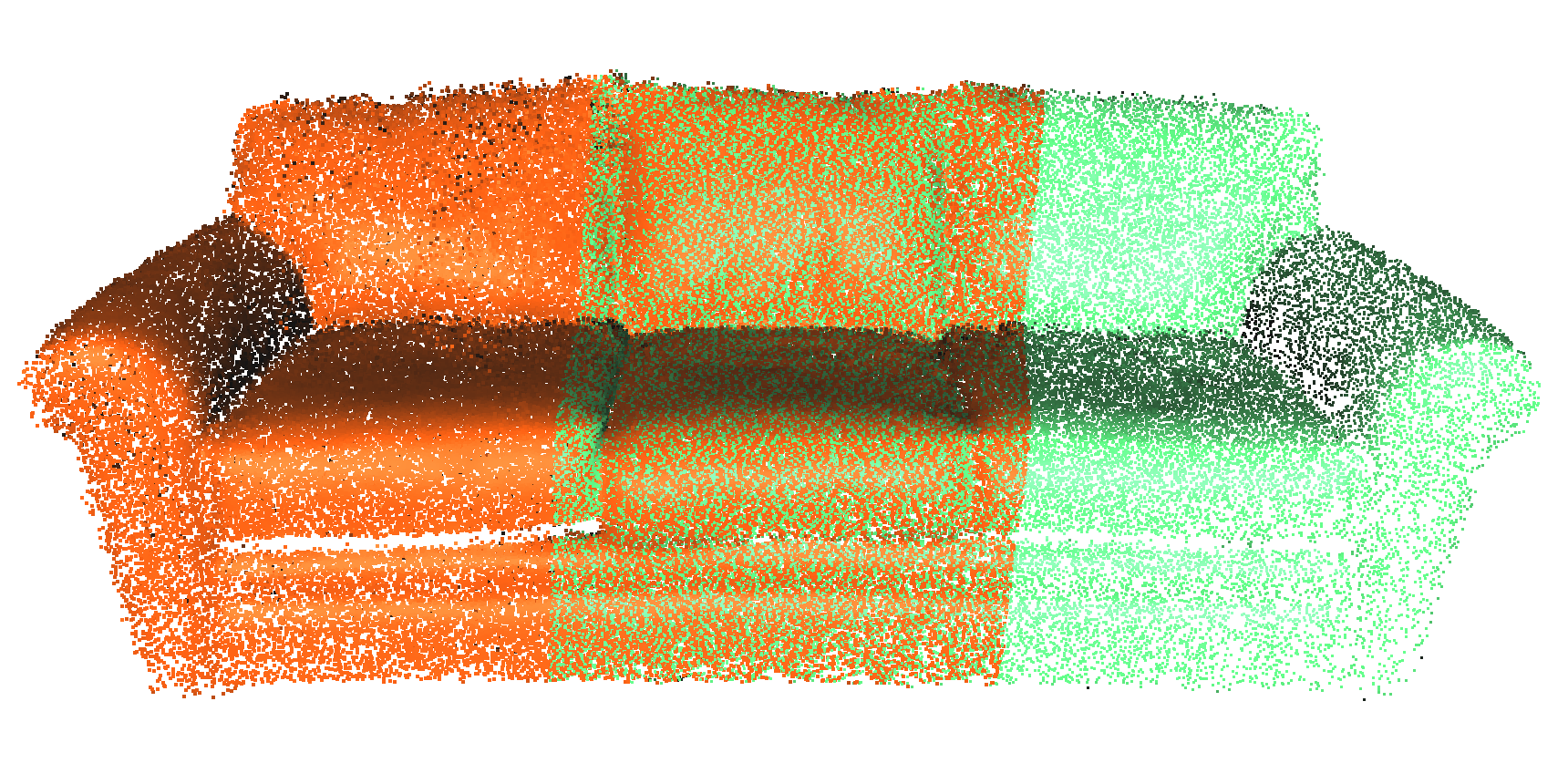}
&
\includegraphics[width=0.5\columnwidth]{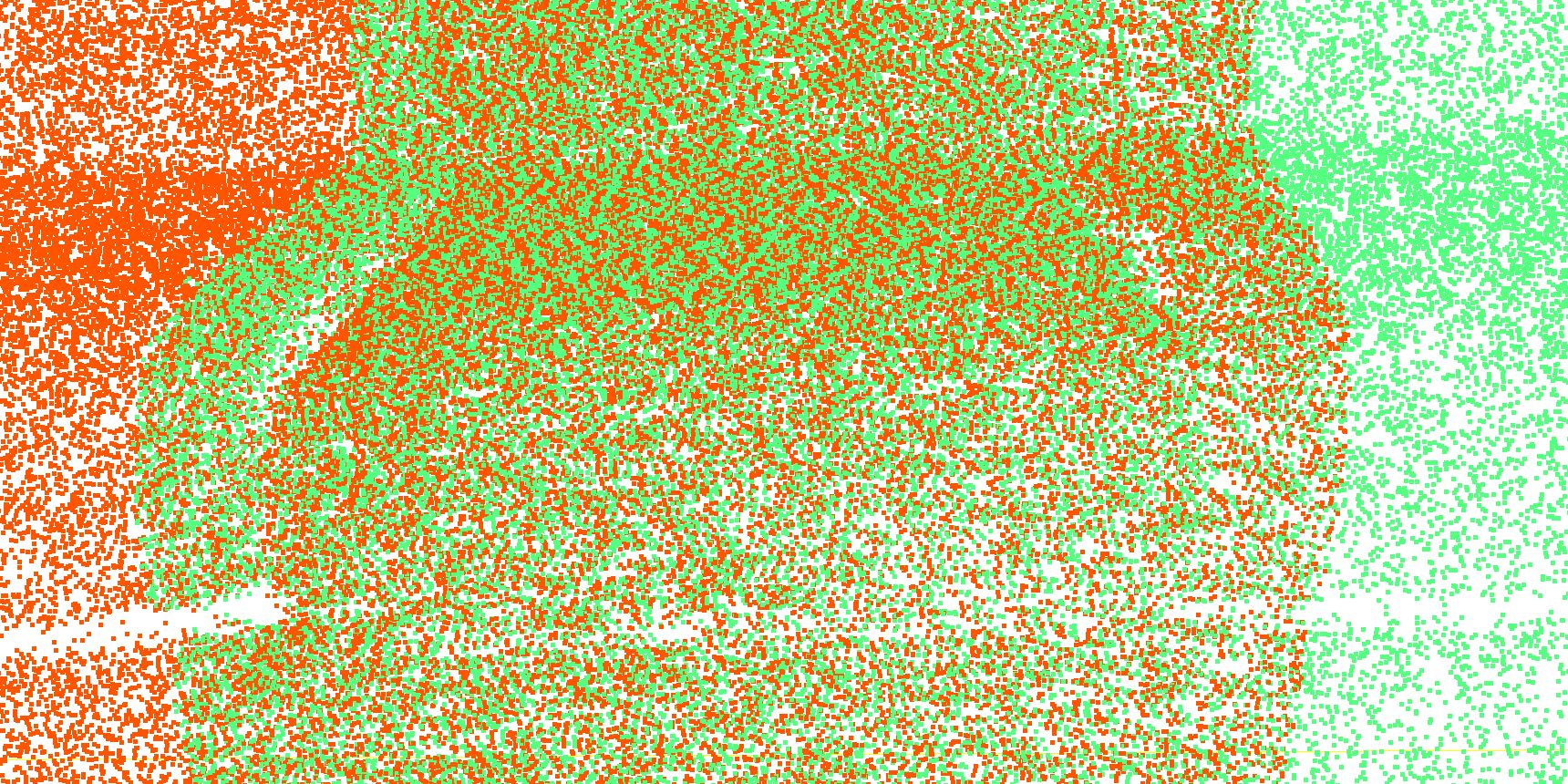}
\\
{\small (d) Uniform resampling ($\downarrow 20$).  } &  {\small  (e) Registered point cloud.  }   &  {\small (f) Details. }
\\
\includegraphics[width=0.5\columnwidth]{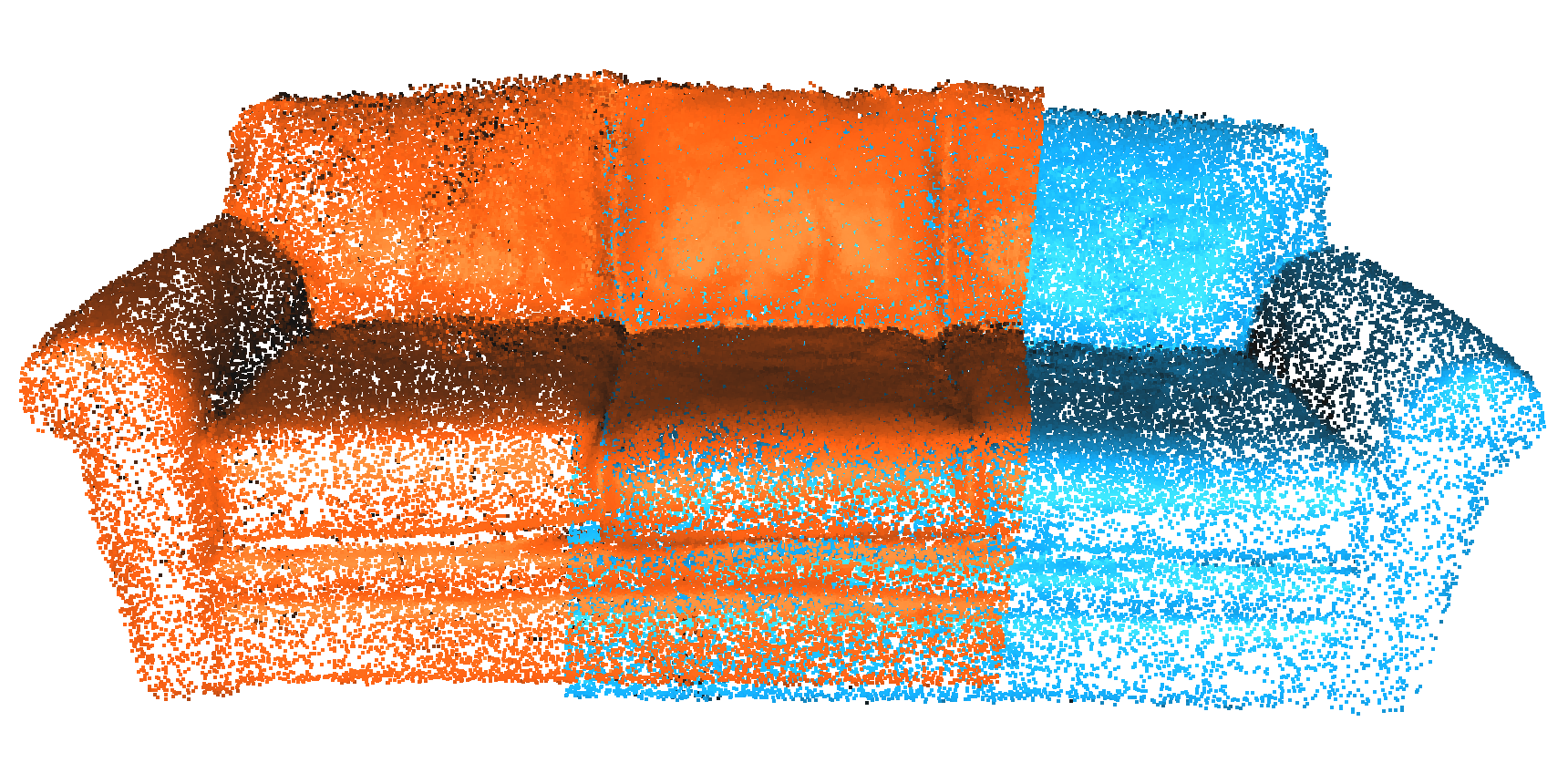}
&
\includegraphics[width=0.5\columnwidth]{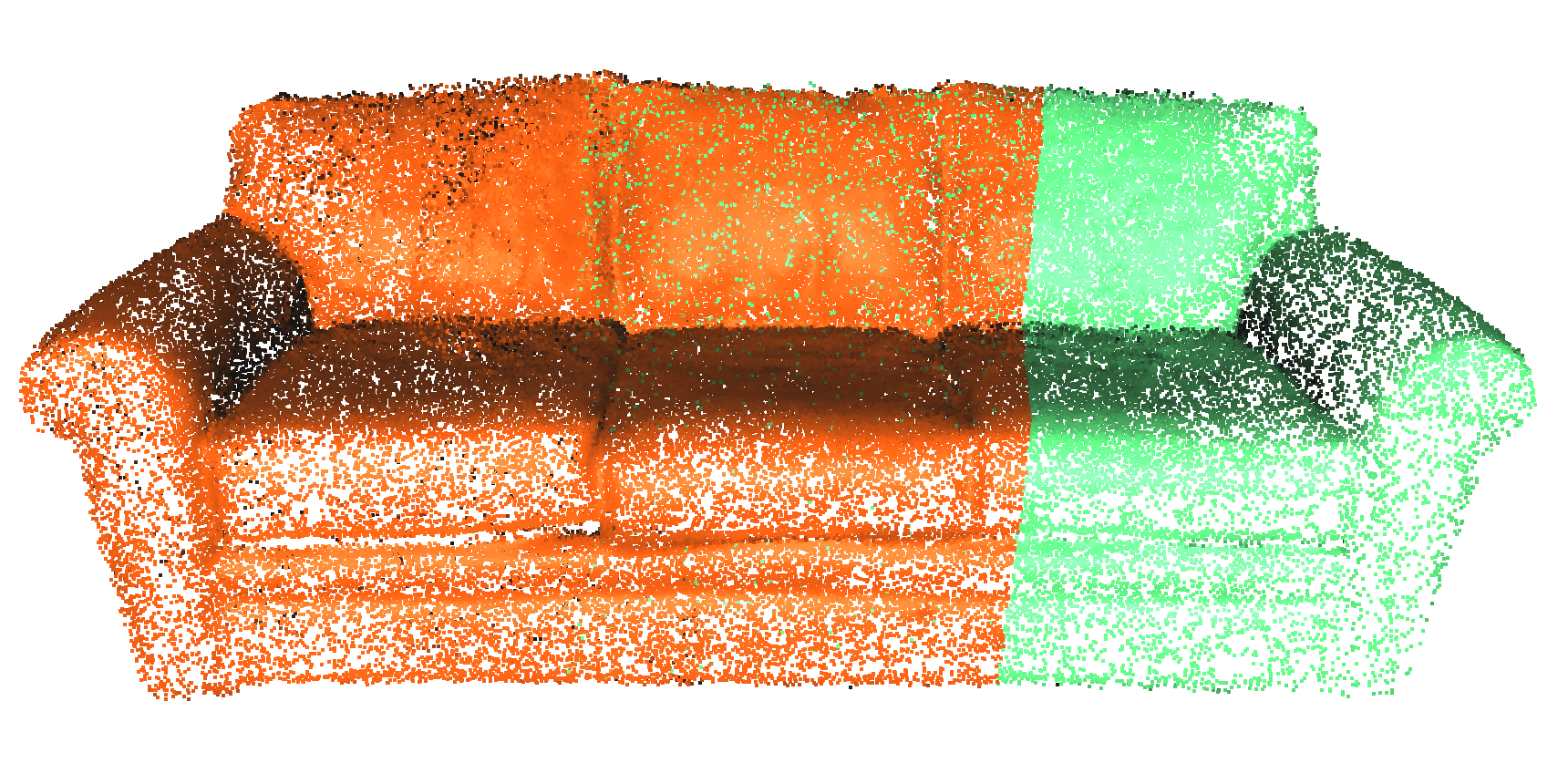}
&
\includegraphics[width=0.5\columnwidth]{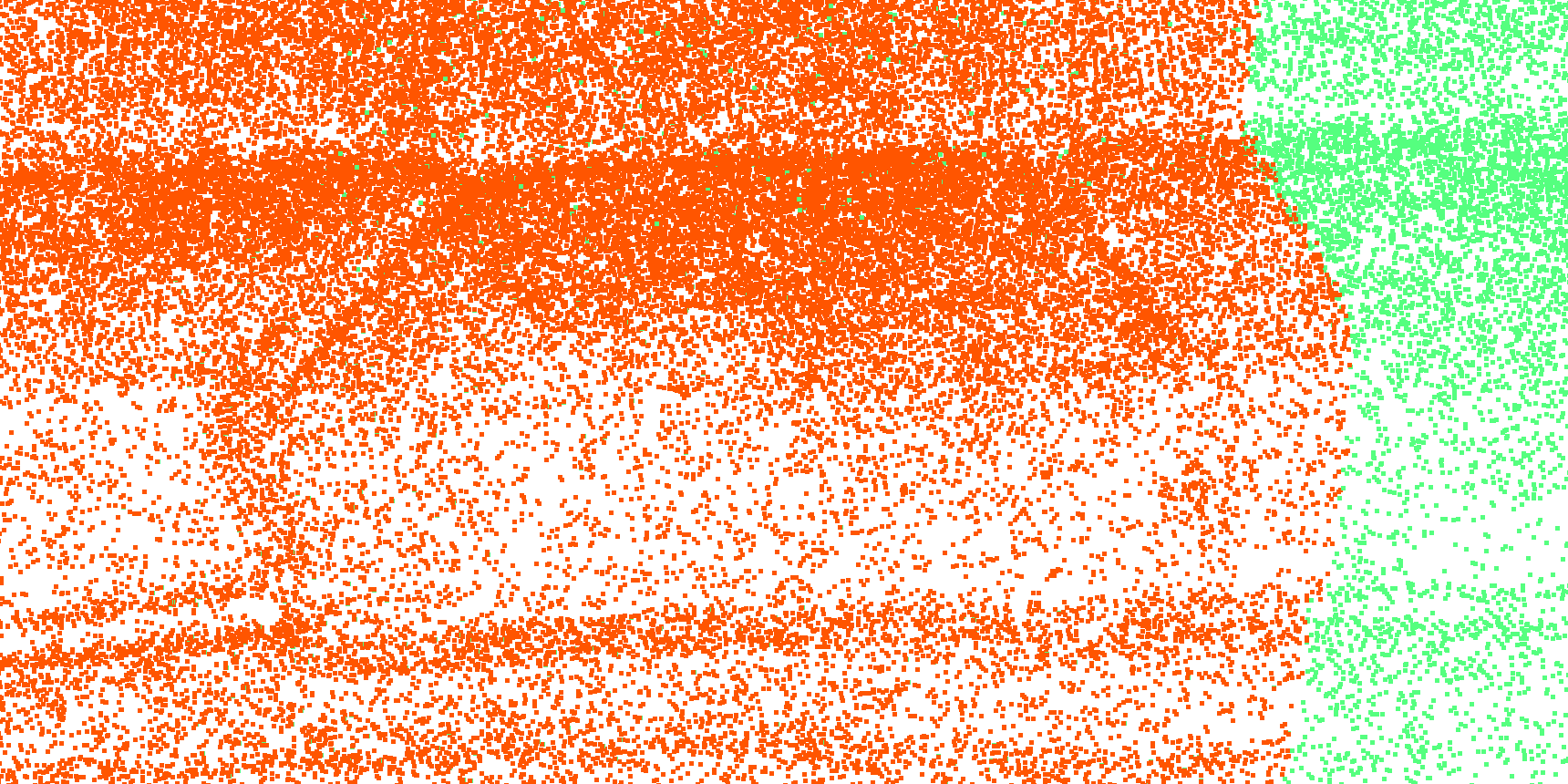}
\\
{\small (g) High-pass graph filtering based } &  {\small  (h) Registered point cloud. }    &  {\small  (i) Details. }   
\\
{\small   graph filtering ($\downarrow 20$). } 
\end{tabular}
  \end{center}
  \caption{\label{fig:sofa_regist}  Accurate registration for sofa. The first row shows the original point cloud; the second row shows the uniformly resampled point cloud; and the third row shows the high-pass graph filtering based resampled   point cloud~\eqref{eq:opt_sample_HH}. The first column shows the point clouds before registration; the second column shows the point clouds after registration; and the second column shows the registration details around the overlapping area. High-pass graph filtering-based resampling provides more precise registration by using fewer points. }
\end{figure*}

\begin{table}[t]
  \footnotesize
  \begin{center}
    \begin{tabular}{@{}llll@{}}
      \toprule
       &  RMSE &  ${\rm Error}_{\rm shift}$  & ${\rm Error}_{\rm rotation}$   \\
      \midrule \addlinespace[1mm]
All points &   $4.22$  & $8.76$  & $2.30 \times 10^{-3}$  \\
Uniform resampling &   $4.27$  & $9.38$  & $3.76  \times 10^{-3}$  \\  
High-pass graph filtering &   ${\bf 1.49}$  & ${\bf 0.01}$ &   ${\bf 4.29  \times 10^{-5}}$ \\    
~based resampling~\eqref{eq:opt_sample_HH} && \\
      \bottomrule
    \end{tabular}
  \end{center}
  \caption{\label{table:sofa_regist} Proposed high-pass graph filtering based resampling strategy provides an accurate registration for a sofa. Best results are marked in bold. High-pass graph filtering based resampling chooses key points and provides the best registration performance. }
  \vspace{-3mm}
\end{table}

In this section, we apply the proposed resampling strategies to  accurate registration.
In this task, we use the proposed resampling strategy~\eqref{eq:opt_sample_HH} to make two point clouds registered  efficiently and accurately.

Figure~\ref{fig:sofa_regist} (a) shows a  point cloud of a sofa, which contains $1,204,055$ points collected from a Kinect based SLAM system~\cite{TaguchiJRF:13}. As shown in Figure~\ref{fig:sofa_regist} (a), we split  the original point cloud into two overlapping point clouds marked in red and blue, respectively. We intentionally shift and rotate the red part. The task is to invert the process and retrieve the shift and rotation. We use the iterative closest point (ICP) algorithm  to register two point clouds, which is a standard algorithm to rotate and shift different scans into a consistent coordinate frame~\cite{Besl:92}. The ICP algorithm iteratively revises the rigid body transformation (combination of shift and rotation) needed to minimize the distance from the source to the reference point cloud. Figures~\ref{fig:sofa_regist} (b) and (c) show the registered sofa and the details of the overlapping part after registration, respectively. We see that the registration process recovers the overall structure of the original point cloud, but still leaves some mismatch in a detailed level.

Since it is inefficient to register two large-scale point clouds, we want to resample a subset of 3D points from each point cloud and implement registration. We will compare the registration performance between 
uniformly resampled point cloud and high-pass graph filtering based resampled point cloud. Note that  high-pass graph filtering based resampling can enhance the contours and key points. Figures~\ref{fig:sofa_regist} (d) and (g) show the 
resampled point clouds based on uniform resampling and high-pass graph filtering based  resampling, respectively. Two resampled versions have the same number of points, which is $5\%$ of points in the original point cloud. We see that Figures~\ref{fig:sofa_regist} (g) shows more contours than Figures~\ref{fig:sofa_regist} (d).  Based on the uniformly resampled version Figures~\ref{fig:sofa_regist} (d), Figures~\ref{fig:sofa_regist} (e) and (f) show the  registered sofa  and the details of the overlapping part after registration, respectively.  Based on the contour-enhanced resampled version Figures~\ref{fig:sofa_regist} (g), Figures~\ref{fig:sofa_regist} (h) and (i) show the  registered sofa  and the details of the overlapping part after registration, respectively. We see that the registration based on high-pass graph filtering based resampling precisely recovers the original point cloud, even in a detailed level. The intuition is that the high-pass graph filters enhance the contours, which make sharper match between the sources and targets, and thus the registration becomes easier.

The quantitative results are shown in Table~\ref{table:sofa_regist}, where the first column shows the root mean square error (RMSE); the second column shows the shift error; The third column shows the rotation error. Specially,
$ {\rm RMSE}  =   \sqrt{ \sum_{i=1}^N \min_{j = 1, \ldots, N}  \left\| \widehat{\x}_i - \x_j \right\|_2^2}$, ${\rm Error}_{\rm shift}  = \left\| \widehat{\a} - \a \right\|_2$ and ${\rm Error}_{\rm rotation} =  \left\| \widehat{\RR} - \RR \right\|_{\rm Frobenius}$, 
where $\widehat{\x}_i$, $\widehat{\a}$ and $\widehat{\RR}$ are the 3D coordinates of the $i$th point, recovered shift vector and recovered rotation matrix after registration, respectively;  $\x_i$, $\a$ and $\RR$ are the ground-truth 3D coordinates of the $i$th point, ground-truth shift vector and ground-truth recovered rotation matrix. We see that high-pass graph filtering based resampled point cloud uses $20$-times fewer points and achieves even better results than using all the points. The shift and rotation errors of using high-pass graph filtering based resampling are significantly smaller than those of using all the points or using uniform resampling.

\section{Conclusions}
\label{sec:conclusions}
In this paper, we proposed a resampling framework to select a subset of points to extract application-dependent features and reduce the subsequent computation in a large-scale point cloud. We formulated an optimization problem to obtain the optimal resampling distribution, which is also guaranteed to be shift/rotation/scale invariant.  We then specified the feature extraction operator to be a graph filter and studied the resampling strategies based on all-pass, low-pass and high-pass graph filtering. A surface reconstruction system based on graph filter banks was introduced to compress 3D point clouds. Three applications, including large-scale visualization, accurate registration and robust shape modeling, were presented to validate the effectiveness and efficiency of the proposed resampling methods. This work also pointed out many possible future directions of 3D point cloud processing, such as efficient 3D point cloud compression system based on graph filter banks, surface reconstruction based on arbitrary graphs, robust metric to evaluate the quality of a 3D point cloud.

% -------------------------------------------------------------------------
\bibliographystyle{IEEEbib}
%\bibliography{bibl_jelena}
\bibliography{refs}

\appendix
\section{Appendices}
\subsection{Proof of Lemma~\ref{lem:unbias}}
\label{sec:app1}
\begin{proof}
For the nonweighted version, we have
\begin{eqnarray*} 
&& \mathbb{E}_{\Psi \sim \pi} \left( \Psi^T \Psi f(\X) \right)_i
 \ = \ \mathbb{E}_{\M}  \left( \sum_{\M_j \in \M} f_{\M_j} (\X) \delta_{\M_j = i}  \right)
\\
& \stackrel{(a)}{=} & M  \mathbb{E}_{\ell}  \left( f_{\ell} (\X) \delta_{\ell = i}  \right)
\ = \ M \sum_{\ell = 1}^N  f_{\ell} (\X)\pi_{\ell} \delta_{\ell = i}  
\\
& = & M \pi_i f_i(\X). 
\end{eqnarray*}
where $\left( \Psi^T \Psi f(\X) \right)_i \in \R^K$ is the $i$th row of $\Psi^T \Psi f(\X)$, $f_i(\X) \in \R^K$ is the $i$th row of $f(\X)$, $\delta_{i}$ denotes an indicator function and equality (a) follows from the independent and identically distributed random sampling.

For the reweighted  version, we have
\begin{eqnarray*} 
&& \mathbb{E}_{\Psi \sim \pi}  \left( \Ss \Psi^T \Psi f(\X) \right)_i
 \\
& = & \mathbb{E}_{\M}  \left( \sum_{\M_j \in \M}  \Ss_{\M_j, \M_j} f_{\M_j} (\X)  \delta_{\M_j = i}  \right)
\\
& = & M  \mathbb{E}_{\ell}  \left( \frac{1}{M \pi_{l}} f_{\ell} (\X)  \delta_{\ell = i}  \right)
\ = \ M \sum_{\ell = 1}^N \frac{1}{M \pi_{\ell}} f_{\ell} (\X)  \pi_{\ell}  \delta_{\ell = i} \\
& = & f_{i} (\X) .
\end{eqnarray*}
\end{proof}

\subsection{Proof of Theorem~\ref{thm:MSE}}
\label{sec:app2}
\begin{proof}
We first split the error into the bias term and the variance term,
\begin{eqnarray*}
&& \mathbb{E}_{\Psi \sim \pi} \left\|  \Ss \Psi^T \Psi  f(\X) -  f(\X) \right\|_2^2
\\
& = & \left\| \mathbb{E}_{\Psi \sim \pi} \left( \Ss \Psi^T \Psi  f(\X) \right) -  f(\X) \right\|_2^2  
\\
&& +\mathbb{E}_{\Psi \sim \pi} \left\| \Ss \Psi^T \Psi  f(\X) - \mathbb{E}_{\Psi \sim \pi} \left(  \Ss \Psi^T \Psi  f(\X)  \right) \right\|_2^2,
\end{eqnarray*}
where the first term is bias and the second term is variance. Lemma~\eqref{lem:unbias} shows that the bias term is zero. So, we only need to bound the variance term.

For each element in the variance term, we have
\begin{eqnarray*} 
&& \mathbb{E}_{\Psi \sim \pi} \left\| \left( \Ss \Psi^T \Psi  f(\X) \right)_i - \mathbb{E} \left( \Ss \Psi^T \Psi  f(\X) \right)_i \right\|^2 
\\
& = & \mathbb{E}_{\M}  \Bigg[ \left( \sum_{\M_j \in \M} \Ss_{\M_j, \M_j} f_{\M_j} (\X)  \delta_{\M_j = i}  - f_i (\X) \right)^T
\\
&&~~\left( \sum_{\M_{j'} \in \M} \Ss_{\M_{j'}, \M_{j'}} f_{\M_{j'}} (\X)  \delta_{\M_{j'} = i}  - f_i (\X) \right) \Bigg]
\\
& = & \mathbb{E}_{\M} \bigg( \sum_{\M_j, \M_{j'}  \in \M} \Ss_{\M_j, \M_j} \Ss_{\M_{j'}, \M_{j'}} f_{\M_j} (\X)^T   f_{\M_{j'}} (\X)
\\
&& ~~~\delta_{\M_j = i} \delta_{\M_{j'} = i}  \bigg)  - f_i (\X)^T f_i (\X)
\\
& = & M^2  \mathbb{E}_{\ell}   \Ss^2_{\ell, \ell} f_{\ell} (\X)^T f_{\ell} (\X) \delta_{\ell = i} - f_i (\X)^T f_i (\X)
\\
& = &
M^2  \sum_{\ell = 1}^N \frac{ f_{\ell} (\X)^T f_{\ell} (\X) }{M^2 \pi_{\ell}^2} \pi_{\ell} \delta_{\ell = i} - f_i (\X)^T f_i (\X)
\\
& = & \left( \frac{1}{ \pi_{i} } - 1 \right) f_i (\X)^T f_i (\X).
\end{eqnarray*}
We finally combine all the elements and obtain~\eqref{eq:MSE}.
\end{proof}

\subsection{Proof of Theorem~\ref{thm:MSE_variant}}
\label{sec:app3}
\begin{proof}
Based on Theorem~\ref{thm:MSE}, we have
\begin{eqnarray*}
&& \mathbb{E}_{\Psi \sim \pi} \left(  D_{f} (\Psi) \right) 
\\ 
& = &   \mathbb{E}_{\Psi \sim \pi} \max_{\X'_c: \left\|  \X'_c \right\|_2 = c }  \left\| 
  \left( \Ss \Psi^T \Psi - \Id \right)  f \left(
  \begin{bmatrix}
  \X'_c & \Xo
\end{bmatrix} \right) \right\|_F^2
\\ 
& = &   \mathbb{E}_{\Psi \sim \pi} \max_{\X'_c: \left\|  \X'_c \right\|_2 = c }  \left\| 
  \left( \Ss \Psi^T \Psi - \Id \right)  \F
  \begin{bmatrix}
  \X'_c & \Xo
\end{bmatrix}  \right\|_F^2
\\
& = &   \mathbb{E}_{\Psi \sim \pi} \bigg( \max_{\X'_c: \left\|  \X'_c \right\|_2 = c }  \left\| 
  \left( \Ss \Psi^T \Psi - \Id \right)  \F  \X'_c  \right\|_F^2 
\\
&& + ~  \left\| 
  \left(  \Ss \Psi^T \Psi - \Id \right)  \F  \Xo  \right\|_F^2   \bigg)
  \\
& = &   \mathbb{E}_{\Psi \sim \pi} \bigg( c^2 \left\| 
 \left( \Ss \Psi^T \Psi - \Id \right)  \F   \right\|_F^2 
\\
&& + ~  \left\| 
  \left( \Ss \Psi^T \Psi - \Id \right)  \F  \Xo  \right\|_F^2   \bigg)
  \\
  & = &  c^2 {\rm Tr} \left( \F \Q  \F^T \right) +  {\rm Tr} \left( \F \Xo \Q  (\F\Xo)^T \right).
\end{eqnarray*}
\end{proof}

\subsection{Proof of Theorem~\ref{thm:opt_sample_invariant}}
\label{sec:app4}
\begin{proof}
The optimal resampling strategy is the solution of the following optimization problem,
\begin{eqnarray}
\label{eq:opt_proof}
&&	\min_{\pi} ~\mathbb{E}_{\Psi \sim \pi} \left(  D_{f(\X)} (\Psi) \right)
	\\
	\nonumber
&& {\rm subject~to~}  \sum_{i=1}^N \pi_i = 1,~~~  \pi_i \geq 0.
\end{eqnarray}

The corresponding Lagrange function is 
\begin{eqnarray*}
&&  L(\pi_i, \lambda, \mu)   \\
& = & 
\mathbb{E}_{\Psi \sim \pi} \left(  D_{f(\X)} (\Psi) \right) + \lambda \left( \sum_{ i = 1}^N  \pi_i - 1 \right) + \sum_{ i = 1}^N \mu_i \pi_i
\\
& = &  \sum_{ i = 1}^N \left( \frac{1}{ \pi_{i} } - 1 \right) \left\| f_i(\X) \right\|_2^2 + \lambda \left( \sum_{ i = 1}^N  \pi_i - 1 \right) + \sum_{ i = 1}^N \mu_i \pi_i,
\end{eqnarray*}
where the equality follows from Theorem~\ref{thm:MSE}. The derivative to $\pi_i$ is 
\begin{equation}
\frac{\partial L}{ \partial \pi_i}  \ = \
- \frac{1}{\pi_i^2} \left\| f_i(\X) \right\|_2^2 + \lambda + \mu_i.
\end{equation}
By setting its derivative to zero, we have 
\begin{equation*}
 \pi_i =   \frac{ \left\| f_i(\X) \right\|_2 }{ \sqrt{\lambda + \mu_i} }.
\end{equation*}
Due to the complementary slackness, we have
$$
\mu_i \pi_i \ = \  \frac{ \mu_i \left\| f_i(\X) \right\|_2  } { \sqrt{\lambda + \mu_i } } \ = \ 0.
$$ Thus, either $\mu_i$ or $\left\| f_i(\X) \right\|_2 $ is zero. In both cases,  $\pi_i \propto \left\| f_i(\X) \right\|_2 $.
\end{proof}

\subsection{Proof of Theorem~\ref{thm:opt_sample_variant}}
\label{sec:app5}
\begin{proof}
The optimal resampling strategy is the solution of the following optimization problem,
\begin{eqnarray}
&&	\min_{\pi} ~\mathbb{E}_{\Psi \sim \pi}  \left(  D_{f} (\Psi) \right)
	\\
	\nonumber
&& {\rm subject~to~}  \sum_{i=1}^N \pi_i = 1,~~~  \pi_i \geq 0.
\end{eqnarray}

The corresponding Lagrange function is 
\begin{eqnarray*}
&&  L(\pi_i, \lambda, \mu) 
\\
& = &
\mathbb{E}_{\Psi \sim \pi} \left(  D_{f} (\Psi) \right) + \mu \left( \sum_{ i = 1}^N  \pi_i - 1 \right) + \sum_{ i = 1}^N \mu_i \pi_i
\\
& = & c^2 \sum_{ i = 1}^N \left( \frac{1}{ \pi_{i} } - 1 \right) \left\| \F_i \right\|_2^2 + \sum_{ i = 1}^N \left( \frac{1}{ \pi_{i} } - 1 \right) \left\| \left( \F \Xo \right)_i \right\|_2^2 
\\
&& + \mu \left( \sum_{ i = 1}^N  \pi_i - 1 \right) + \sum_{ i = 1}^N \mu_i \pi_i,
\end{eqnarray*}
where $\F_i$ is the $i$th row of $\F$ and $\left( \F \Xo \right)_i $ is the $i$th row of $\F \Xo $. The derivative to $\pi_i$ is 
\begin{equation*}
\frac{\partial L}{ \partial \pi_i}  \ = \
- \frac{1}{\pi_i^2}  \left(   c^2 \left\| \F_i \right\|_2^2  +   \left\| \left( \F \Xo \right)_i \right\|_2^2  \right)
+ \mu + \mu_i.
\end{equation*}
By setting its derivative to zero, we have 
\begin{equation*}
 \pi_i =   \frac{ \sqrt{  c^2 \left\| \F_i \right\|_2^2  +   \left\| \left( \F \Xo \right)_i \right\|_2^2 } }{ \sqrt{\mu + \mu_i} }.
\end{equation*}
Due to the complementary slackness, we have
$$
\mu_i \pi_i \ = \  \frac{ \mu_i  \sqrt{  c^2 \left\| \F_i \right\|_2^2  +   \left\| \left( \F \Xo \right)_i \right\|_2^2 }  } { \sqrt{\mu + \mu_i } } \ = \ 0.
$$ Thus, either $\mu_i$ or $\left\| f_i(\X) \right\|_2 $ is zero. In both cases,  $\pi_i \propto \sqrt{  c^2 \left\| \F_i \right\|_2^2  +   \left\| \left( \F \Xo \right)_i \right\|_2^2 } $.
\end{proof}

\subsection{Proof of Theorem~\ref{thm:feature_invariant}}
\label{sec:app6}
\begin{proof}
We first show the rotational invariance. Let $\X$ be the 3D coordinates of an original point cloud and $\RR \in \R^{3 \times 3}$ be a rotation matrix. The point cloud after rotating is $\X \RR$. The local variation of $\X \RR$ is
\begin{eqnarray*}
	f_i (\X \RR ) & = & \left\|  \left( h(\Adj) \X \RR \right)_i \right\|_2^2
	\\
	& = &   \left\|  \left( h(\Adj)  \right)_i  \X \RR \right\|_2^2
	\\
	& = &  \left( h(\Adj)  \right)_i  \X \RR  \RR^T \X^T \left( h(\Adj)  \right)_i ^T
	\\
	& \stackrel{(a)}{=} &  \left( h(\Adj)  \right)_i  \X \X^T \left( h(\Adj)  \right)_i ^T
	\\
	& = &  \left\|  \left( h(\Adj) \X \right)_i \right\|_2^2 \ = \ f_i ( \X ),
\end{eqnarray*}
where $\left( h(\Adj)  \right)_i $ is the $i$th row of $h(\Adj)$ and (a) follows from any rotation matrix $\RR$ is orthonormal.

We next show the shift variance.  Let $\a \in \R^3$ be the shift and the point cloud after shifting is $\X + \one \a^T$. The local variation of $\X + \one \a^T$ is
\begin{eqnarray*}
	f_i ( \X  + \one \a^T ) & = & \left\|  \bigg( h(\Adj)  \left( \X  + \one \a^T \right)  \bigg)_i \right\|_2^2
	\\
	& = & \left\|  \bigg( h(\Adj) \X  \bigg)_i  +   \bigg( h(\Adj) \one \a^T   \bigg)_i \right\|_2^2
\end{eqnarray*}
Thus, $f_i ( \X  + \one \a^T ) = f_i ( \X )$ only when $h(\Adj) \one = \bold{0} \in \R^N$.
\end{proof}

\end{document}